%% file: ms.tex
\documentclass[sigconf]{acmart}

\usepackage{booktabs} 
\usepackage{wrapfig}
\usepackage{bbm}
\usepackage{xspace}
\usepackage{amssymb}
\usepackage{pifont}
\usepackage{algorithm}
\usepackage{algorithmic}
\usepackage{footmisc}
 \usepackage{multicol}

\usepackage{hyperref}
\usepackage{xcolor}
\usepackage{paralist}
\usepackage{subcaption}
\usepackage{enumitem}
\usepackage{color}
\input{dfn}

\hypersetup{
	colorlinks,
	linkcolor={red},
	citecolor={black},
	urlcolor={blue}
}
\setlist[itemize]{leftmargin=0.25in}

\copyrightyear{2018} 
\acmYear{2018} 
\setcopyright{acmcopyright}
\acmConference[CIKM '18]{2018 ACM Conference on Information and Knowledge Management}{October 22--26, 2018}{Torino, Italy}
\acmBooktitle{2018 ACM Conference on Information and Knowledge Management (CIKM'18), October 22--26, 2018, Torino, Italy}
\acmPrice{15.00}
\acmDOI{10.1145/XXXXXX.XXXXXX}
\acmISBN{978-1-4503-6014-2/18/10} 
    
\begin{document}
	


\title{A Quest for Structure: Jointly Learning the Graph Structure and Semi-Supervised Classification}

\author{Xuan Wu}
\authornote{X. Wu \& L. Zhao contributed equally to the paper}
\affiliation{%
    \institution{School of Computer Science\\
        Carnegie Mellon University}
}
\email{xuanw1@andrew.cmu.edu}

\author{Lingxiao Zhao}
\authornotemark[1]
\affiliation{%
    \institution{H. John Heinz III College\\
        Carnegie Mellon University}
}
\email{lingxiao@cmu.edu}

\author{Leman Akoglu}
\affiliation{%
	\institution{H. John Heinz III College\\
		Carnegie Mellon University}
}
\email{lakoglu@andrew.cmu.edu}

\renewcommand{\shorttitle}{Joint Graph Structure and Semi-Supervised Learning}

\begin{abstract}

\input{00abstract.tex}
\end{abstract}

%
%


\maketitle

\section{Introduction}
\label{01intro}
\input{01intro}

\section{Related Work}
\label{04related}
\input{04related}
\section{Preliminaries and Background}
\label{02prelims}
\input{02prelims}

\section{Proposed Method: \method}
\label{03proposed}
\input{03proposed}

\input{03validation}

\input{03parallel}

\section{Evaluation}
\label{03experiments}
\input{03experiments}

\section{Conclusion}
\label{05conclusion}
\input{05conclusion}


%
%
\begin{acks}
{\small This research is sponsored by NSF CAREER 1452425 and IIS 1408287. Any conclusions in this material are of the authors and do not necessarily reflect the views of the funding parties.}
\end{acks}

\bibliographystyle{ACM-Reference-Format}
\bibliography{refs} 

\end{document}

%% file: dfn.tex
\newcommand\footnoteref[1]{\protected@xdef\@thefnmark{\ref{#1}}\@footnotemark}
\newcommand{\method}{{\sc PG-learn}}
\newcommand{\grad}{{\sc Gradient}}

\newcommand{\grid}{{\em Grid}}
\newcommand{\randd}{{\em Rand$_d$}}
\newcommand{\minent}{{\em MinEnt}}
\newcommand{\metric}{{\em IDML}}
\newcommand{\aew}{{\em AEW}}

\newcommand{\coil}{\textsc{COIL}}
\newcommand{\usps}{\textsc{USPS}}
\newcommand{\umist}{\textsc{UMIST}}
\newcommand{\mnist}{\textsc{MNIST}}
\newcommand{\yale}{\textsc{Yale}}

\newcommand{\s}{$^\blacktriangle$}
\newcommand{\hs}{$^\triangle$}

\newcommand{\cbit}{\begin{compactitem}}
	\newcommand{\ceit}{\end{compactitem}}
\newcommand{\cben}{\begin{compactenum}}
	\newcommand{\ceen}{\end{compactenum}}

\definecolor{OliveGreen}{rgb}{0,0.6,0}

\newcommand{\bit}{\begin{itemize}}
	\newcommand{\eit}{\end{itemize}}
\newcommand{\ben}{\begin{enumerate}}
	\newcommand{\een}{\end{enumerate}}
\newcommand{\beq}{\begin{equation}}
\newcommand{\eeq}{\end{equation}}

\newcommand{\mD}{\mathcal{D}}
\newcommand{\mG}{\mathcal{G}}
\newcommand{\mK}{\mathcal{K}}
\newcommand{\mL}{\mathcal{L}}
\newcommand{\mV}{\mathcal{V}}
\newcommand{\mC}{\mathcal{C}}

\newcommand{\mR}{\mathbb{R}}
\newcommand{\mN}{\mathbb{N}}
\newcommand{\mB}{\mathbb{B}}

\newcommand{\bx}{\boldsymbol{x}}
\newcommand{\ba}{\boldsymbol{a}}
\newcommand{\bY}{\boldsymbol{Y}}
\newcommand{\bF}{\boldsymbol{F}}
\newcommand{\bW}{\boldsymbol{W}}
\newcommand{\bA}{\boldsymbol{A}}
\newcommand{\bL}{\boldsymbol{L}}
\newcommand{\bI}{\boldsymbol{I}}
\newcommand{\bD}{\boldsymbol{D}}
\newcommand{\bP}{\boldsymbol{P}}
\newcommand{\bX}{\boldsymbol{X}}
\newcommand{\bone}{\boldsymbol{1}}

\newcommand{\hide}[1]{}

%% file: 00abstract.tex
Semi-supervised learning (SSL) is effectively used for numerous classification problems,
thanks to its ability to make use of abundant unlabeled data. 
The main assumption of various SSL algorithms is that the nearby points  on the data manifold are likely to share a label.
Graph-based SSL constructs a graph from point-cloud data as an approximation to the underlying manifold, followed by label inference.
It is no surprise that the quality of the constructed graph in capturing the essential structure of the data  is critical to the accuracy of the subsequent inference step \cite{conf/pkdd/SousaRB13}.

How should one construct a graph from the 
input point-cloud data for graph-based SSL?
In this work we introduce a new, parallel graph learning framework (called \method) for the graph construction step of SSL. 
Our solution has two main ingredients: 
(1) a gradient-based optimization of the edge weights (more specifically, different kernel bandwidths in each dimension) based on a validation loss function,
and (2) a parallel hyperparameter search algorithm with an adaptive resource allocation scheme.
In essence, (1) allows us to search around a (random) initial hyperparameter configuration for a better one with lower validation loss. Since the search space of hyperparameters is huge for high-dimensional problems, (2) empowers our gradient-based search to go through as many different initial configurations as possible, where runs for relatively unpromising starting configurations are terminated early to allocate the time for others.
As such, \method~ is a carefully-designed hybrid of random and adaptive search.
Through experiments on multi-class classification problems, we show
that \method~ significantly outperforms a {variety} of existing graph construction schemes in accuracy (per fixed time budget for hyperparameter tuning), and scales more effectively to high dimensional problems.


%% file: 01intro.tex

Graph-based semi-supervised learning (SSL) algorithms, based on graph min-cuts \cite{conf/icml/BlumC01}, local and global consistency \cite{conf/nips/ZhouBLWS03}, and harmonic energy minimization \cite{zhuGhahramLaff03semisup}, have been used widely for classification and regression problems.
These employ the manifold assumption to take advantage of the unlabeled data, which dictates the label (or value) function to change smoothly on the data manifold. 

The data manifold is modeled by a graph structure. In some cases, this graph is explicit; for example, explicit social network connections between individuals have been used in predicting their political orientation \cite{conf/socialcom/ConoverGRFM11}, age \cite{conf/www/PerozziS15}, income \cite{conf/acl/FlekovaPU16}, occupation \cite{conf/acl/Preotiuc-Pietro15}, etc.
In others (e.g., image classification), the data is in (feature) vector form, where a graph is to be constructed from point-cloud data. 
In this graph, nodes correspond to labeled and unlabeled data points and edge weights encode pairwise similarities.
Often, some graph sparsification scheme is also used to ensure that the SSL algorithm runs efficiently.
Then, labeling is done in such a way that instances connected by large weights are assigned similar labels.


In essence, graph-based SSL for non-graph data consists of two steps: (1) graph construction, and (2) label inference. It is well-understood in areas such as clustering and  outlier detection  that the choice of the similarity measure 
has considerable effect on the outcomes. 
Specifically,  Maier et al.  demonstrate the critical influence of graph construction on graph-based clustering \cite{conf/nips/MaierLH08}.
Graph-based SSL is no exception. A similar study by de Sousa et al.  find that ``SSL algorithms are strongly affected by the graph sparsification parameter value and the choice of the adjacency graph construction and weighted matrix generation methods'' \cite{conf/pkdd/SousaRB13}.

Interestingly, however, the (1)st step---graph construction for SSL---is notably under-emphasized in the literature as compared to the (2)nd step---label inference algorithms.  
Most practitioners default to using a similarity measure such as radial basis function (RBF), coupled with sparsification by  $\epsilon$-neighborhood (where node pairs only within distance $\epsilon$ are connected) or $k$NN (where each node is connected to its $k$ nearest neighbors).
Hyper-parameters, such as RBF bandwidth $\sigma$ and $\epsilon$ (or $k$),  are then selected by grid search based on cross-validation error.

{There exist some work on graph construction for SSL beyond $\epsilon$- and $k$NN-graphs, which we review in  \S\ref{04related}.
Roughly, related work can be split into unsupervised and supervised techniques. All of them suffer from one or more drawbacks in terms of efficient search, scalability, and graph quality for the given SSL task. More specifically, unsupervised methods do not leverage the available labeled data for learning the graph.
On the supervised side, most methods are not task-driven, that is, they do not take into account the given SSL task to evaluate graph quality and guide the graph construction, or do not effectively scale to high dimensional data in terms of both runtime and memory. 
Most importantly, 
the graph learning problem is typically non-convex and comes with a prohibitively large search space that should be explored strategically, which is not addressed by existing work.

In this work, we address the problem of graph (structure) learning for SSL, suitable and scalable to high dimensional problems.
We set out to perform the graph construction and label inference steps of semi-supervised learning \textit{simultaneously}.
To this end, we learn different RBF bandwidths $\sigma_{1:d}$ for each dimension, by adaptively minimizing a function of validation loss using (iterative) gradient descent.
In essence, these different bandwidths become model hyperparameters that provide a more general edge weighting function, which in turn can more flexibly capture the underlying data manifold.
Moreover, it is a form of feature selection/importance learning that becomes essential in high dimensions with noisy features. 
On the other hand,  this introduces a scale problem for high-dimensional datasets as we discussed earlier, that is, a large search space with numerous hyperparameters to tune.

Our solution to the scale problem is a Parallel Graph Learning algorithm, called \method, which is a hybrid of random search and adaptive search. It is motivated by  the successive halving strategy \cite{DBLP:journals/corr/LiJDRT16}, which has been recently proposed for efficient hyperparameter optimization for \textit{iterative} machine learning algorithms.
The idea is to start with $N$ hyperparameter settings in parallel threads, adaptively update them for a period of time (in our case, via gradient iterations), discard the worst $N/2$ (or some other fraction) based on validation error, and repeat this scheme for a number of rounds until the time budget is exhausted. In our work, we utilize the idle threads whose hyperparameter settings have been discarded by starting new random configurations on them. Using this scheme, our search tries various random initializations but instead of adaptively updating them fully to completion (i.e., gradient descent convergence), it early-quits those whose progress is not promising (relative to others). While promising configurations are allocated more time for adaptive updates, the time saved from those early-terminations are utilized to initiate new initializations, empowering us to efficiently navigate the search space. 

Our main contributions are summarized as follows.
\bit
\item \textbf{Graph learning for SSL:}
We propose an efficient and effective gradient-based graph (structure) learning algorithm, called \method~(for Parallel Graph Learning), which \textit{jointly} optimizes both steps of graph-based SSL: graph construction and label inference. 

\item \textbf{Parallel graph search with adaptive strategy:}
In high dimensions, it becomes critical to effectively explore the (large) search space. To this end, we couple our (1) \textit{iterative/sequential} gradient-based \textit{local} search with (2) a \textit{parallel}, {resource-adaptive}, \textit{random} search scheme.
In this hybrid, the gradient search runs in parallel with different random initializations, the relatively unpromising fraction of which is terminated early to allocate the time for other initializations in the search space. In effect, (2) empowers (1) to explore the search space more efficiently.

\item \textbf{Efficiency and scalability:}
We use tensor-form gradient (which is more compact and efficient), and make full use of the sparsity of $k$NN graph 
to reduce runtime and memory requirements.
Overall, \method~ scales linearly in dimensionality $d$ and log-linearly in number of samples $n$ computationally, while memory complexity is  linear in both $d$ and $n$. 
\eit

Experiments on multi-class classification tasks show that the proposed \method~ significantly outperforms a variety of existing graph construction schemes  in terms of test accuracy per fixed time budget for hyperparameter search, and further tackles high dimensional, noisy problems more effectively. 

{\bf Reproducibility:} The source code can be found at project page {\url{https://pg-learn.github.io/}. All datasets used in experiments are publicly available (See \S\ref{ssec:databaselines}).

%
%
%

%
%
%
%

%% file: 04related.tex


Semi-supervised learning for non-network data consists of two steps: (1) constructing a graph from the input cloud of points, and (2) inferring the labels of unlabeled examples (i.e., classification). The label inference step has been studied and applied widely, with numerous SSL algorithms \cite{conf/icml/BlumC01,journals/jmlr/BelkinNS06,conf/nips/ZhouBLWS03,zhuGhahramLaff03semisup,conf/cvpr/LiuC09}. On the other hand, the graph construction step that precedes inference has relatively less emphasis in the literature, despite its impact on the downstream inference step.
Our work focuses on this former graph construction step of SSL, as motivated by the findings of de Sousa et al. \cite{conf/pkdd/SousaRB13} and Zhu \cite{zhu05survey}, which show the critical impact of graph construction on clustering and classification problems.

Among existing work, a group of graph construction methods are unsupervised, which do not leverage any information from the labeled data. The most typical ones include similarity-based methods such as $\epsilon$-neighborhood graphs, $k$ nearest neighbor ($k$NN) graphs  and mutual variants.
Jebara et al. introduced the b-matching method \cite{Jebara:2009} toward a balanced graph in which all nodes have the same degree.
There are also self-representation based approaches, like locally linear embedding (LLE) \cite{roweis2000ndr}, low-rank representation (LRR) \cite{Liu:2010:RSS:3104322.3104407}, and variants \cite{Daitch:2009,journals/tip/ChengYYFH10,journals/tcyb/ZhangHHL14}, which model each instance to be a weighted linear combination of other instances where nodes with non-zero coefficients are connected. Karasuyama and Mamitsuka \cite{journals/ml/KarasuyamaM17} extend the LLE idea  by restricting the regression coefficients (i.e., edge weights) to be derived from Gaussian kernels that forces the weights to be positive and greatly reduces the number of free parameters. 
Zhu et al. \cite{zhuGhahramLaff03semisup} proposed to learn different $\sigma_d$ hyperparameters per dimension for the Gaussian kernel by minimizing the entropy of the solution on unlabeled instances via gradient descent. 
Wang et al. \cite{wang2016scalable} focused on the scalability of graph construction by improving Anchor Graph Regularization algorithms, which transform the similarity among samples into similarity between samples and anchor points. 



A second group of graph construction methods are supervised and make use the of labeled data in their optimization. 
Dhillon et al. \cite{dhillon_acl10} proposed a distance metric learning approach within a self-learning scheme to learn the similarity function. However, metric learning uses expensive SDP solvers that do not scale to very large dimensions.
Rohban and Rabiee \cite{journals/pr/RohbanR12} proposed a supervised graph construction approach, showing that under certain  manifold sampling rates, the optimal neighborhood graph is a subgraph of the $k$NN graph family.
Similar to \cite{zhuGhahramLaff03semisup}, 
Zhang and Lee \cite{conf/nips/ZhangL06} also  tune $\sigma_d$'s for different dimensions using a gradient based method, where they minimize the leave-one-out prediction error on labeled data points. Their loss function, however, is specific to the binary classification problems.
Li et al. \cite{li2016graph} proposed a semi-supervised SVM formulation to derive a robust and non-deteriorated SSL by combining multiple graphs together, and it can be used to judge the quality of graphs.
Zhuang et al. \cite{zhuang2017label} incorporated labeling information to graph construction period for self-representation based approach by explicitly enforcing sample can only be represented by samples from the same class.

The above approaches to graph construction have a variety of drawbacks; and typically lack one or more of efficiency, scalability, and graph quality for the given SSL task.
 Specifically, Zhu et al.'s MinEnt \cite{zhuGhahramLaff03semisup} only maximizes confidence over unlabeled samples without using any label information; 
b-matching method \cite{Jebara:2009} only creates a balanced sparse graph which is not a graph \textit{learning} algorithm;
self-representation based methods \cite{roweis2000ndr,Liu:2010:RSS:3104322.3104407,journals/tip/ChengYYFH10,journals/tcyb/ZhangHHL14,journals/ml/KarasuyamaM17}  assume each instance to be a weighted linear combination of other data points and connect those with non-zero coefficients, however such a graph is not necessarily suitable nor optimized specifically for the given SSL task; 
Anchor Graph Regularization \cite{wang2016scalable} only stresses on scalability without considering the graph learning aspect; and several
other graph learning algorithms connected with the SSL task \cite{dhillon_nescai10,conf/nips/ZhangL06} are not scalable in both runtime and memory. 

Our work differs from all existing graph construction algorithms in the following aspects:
(1) \method~ is a gradient-based task-driven graph learning method, which aims to find an optimized graph (evaluated over validation set) for a specific graph-based SSL task; 
(2) \method~ achieves scalability over both dimensionality $d$ and sample size $n$ in terms of runtime and memory. Specifically, it has $O(nd)$ memory complexity and $O(nd+ n\log n)$ computational complexity for each gradient update. 
(3)  Graph learning problem typically has a very large search space with a non-convex optimization objective, where initialization becomes extremely important. To this end,  we design an efficient adaptive search framework outside the core of graph learning. 
This is not explicitly addressed by those prior work, whereas it is one of the key issues we focus on through the ideas of relative performance and 
early-termination.

%% file: 02prelims.tex
\subsection{Notation}
Consider  $\mD := \{(\bx_1,y_1), \ldots, (\bx_l,y_l), \bx_{l+1},\ldots, \bx_{l+u}\}$, a data sample in which the first $l$ examples are labeled, i.e., $\bx_i \in \mR^d$ has label $y_i \in \mN_c$ where 
$c$ is the number of classes and
$\mN_c :=\{p \in \mN^*|1\leq p\leq c\}$. 
Let $u := n-l$ be the number of unlabeled examples
and $\bY \in \mB^{n\times c}$ be a binary label matrix in which $\bY_{ij} = 1$ if and only if $\bx_i$ has label $y_i = j$. 

\sloppy{
The semi-supervised learning task is to assign labels $\{y_{l+1}\ldots,y_{l+u}\}$ to the unlabeled instances. }

\subsection{Graph Construction}

A preliminary step to graph-based semi-supervised learning is the construction of a graph from the point-cloud data.
The graph construction process generates a graph $\mG$ from $\mD$ in which 
each $\bx_i$ is a node of $\mG$.
To generate a  weighted matrix $\bW \in \mR^{n\times n}$ from 
$\mG$, one uses a similarity function $\mK: \mR^d \times \mR^d \rightarrow \mR$ to compute the weights $\bW_{ij}=\mK(\bx_i,\bx_j)$.

A widely used similarity function is the RBF (or Gaussian) kernel,
$
\mK(\bx_i,\bx_j) = \exp(-\|\bx_i - \bx_j\|/(2\sigma^2))
$,
in which $\sigma \in \mR^*_{+}$ is the kernel bandwidth parameter.

To sparsify the graph, two techniques are used most often.
In $\epsilon$-neighborhood ($\epsilon$N) graphs,
there exists an undirected edge between $\bx_i$ and $\bx_j$ 
if and only if $\mK(\bx_i,\bx_j) \geq \epsilon$, where $\epsilon \in \mR^*_{+}$ 
 is a free parameter. 
$\epsilon$ thresholding is prone to generating disconnected or almost-complete graphs 
for an improper value of 
$\epsilon$. 
On the other hand, in the $k$ nearest neighbors ($k$NN) approach,
there exists an undirected edge between $\bx_i$ and $\bx_j$ 
 if either $\bx_i$ or $\bx_j$  is one of the $k$ closest examples to the other.
$k$NN approach has the advantage of being robust to choosing an inappropriate fixed threshold.

In this work, we use a general kernel function
that enables a more flexible graph family, in particular
\begin{equation}
\label{wrbf}
\mK(\bx_i,\bx_j) = \exp \bigg(- \sum_{m=1}^d \frac{(\bx_{im} - \bx_{jm})^2}{\sigma_{m}^2} \bigg) \;,
\end{equation}
where $\bx_{im}$ is the $m^{th}$ component of $\bx_i$.
We  denote $\bW_{ij} = \exp \big(-(\bx_i - \bx_j)^T \bA\; (\bx_i - \bx_j) \big)$,
where $\bA:=diag(\ba)$ is a diagonal matrix with $\bA_{mm}=a_m=1/\sigma_m^2$, that corresponds to
a metric in which different dimensions/features are given different ``weights'', which allows a form of feature selection.\footnote{Setting $\bA$ equal to (i) the identity,
(ii) the (diagonal) variance, or (iii) the covariance matrix would compute similarity 
based on Euclidean, normalized Euclidean, or Mahalanobis distance, respectively.}
In addition, we employ $k$NN graph construction for sparsity.

Our goal is to \textit{learn} both $k$ as well as all the $a_m$'s, 
by means of which we aim to construct a graph that is suitable for the semi-supervised learning task at hand.

\subsection{Graph based Semi-Supervised Learning}

Given the constructed graph $\mG$, a graph-based SSL algorithm uses $\bW$ and the label matrix $\bY$ to generate output matrix $\bF$ by label diffusion in the weighted graph. Note that this paper focuses on the multi-class classification problem, hence $\bF\in \mR^{n\times c}$.

There exist a number of SSL algorithms with various objectives. Perhaps the most widely used ones include
the Gaussian Random Fields algorithm by Zhu et al. \cite{zhuGhahramLaff03semisup},
Laplacian Support Vector Machine algorithm by Belkin et al. \cite{journals/jmlr/BelkinNS06},
and Local and Global Consistency (LGC) algorithm by Zhou et al. \cite{conf/nips/ZhouBLWS03}.

The topic of this paper is how to effectively learn the hyperparameters of graph construction. Therefore, we focus on how the performance of a given recognized SSL algorithm can be improved by means of learning the graph, rather than 
comparing the performance of different semi-supervised or supervised learning algorithms.
To this end, we use the LGC algorithm \cite{conf/nips/ZhouBLWS03}  which we briefly review here.
It is easy to follow the same way to generalize the graph learning ideas introduced in this paper for other popular SSL algorithms, such as Zhu et al.'s
\cite{zhuGhahramLaff03semisup} and Belkin et al.'s \cite{journals/jmlr/BelkinNS06} that have similar objectives to LGC, which we do not pursue further.

The LGC algorithm  solves the optimization problem
\begin{equation}
\label{lgc}
\arg \min_{\bF\in \mR^{n\times c}} tr((\bF-\bY)^T(\bF-\bY) + \alpha \bF^T \bL \bF)\;,
\end{equation}
 where $tr()$ denotes matrix trace,
 $\bL := \bI_n - \bP$ is the normalized graph Laplacian, such that
 $\bI_n$ is the $n$-by-$n$ identity matrix,
  $\bP=\bD^{-1/2}\bW\bD^{-1/2}$, $\bD := diag(\bW\bone_n)$ and $\bone_n$ is the $n$-dimensional all-1's vector.
Taking the derivative w.r.t. $\bF$ and reorganizing the terms, we would get the closed-form solution
$\bF = (\bI_n + \alpha \bL)^{-1}\bY$. 

The solution can also be found without explicitly taking any matrix inverse and instead using the power method \cite{matrixmethods89}, as
\begin{align}
	\label{power}
(\bI + \alpha \bL) \bF  = \bY 
\Rightarrow &\bF + \alpha \bF = \alpha \bP \bF + \bY 
\Rightarrow \bF  = \frac{\alpha}{1+\alpha} \bP \bF + \frac{1}{1+\alpha} \bY \nonumber \\
\Rightarrow &\;\; \bF^{(t+1)}  \leftarrow  \mu \bP \bF^{(t)} + (1-\mu) \bY \;.
\end{align}

\subsection{Problem Statement}

We address the problem of graph (structure) learning for SSL. 
Our goal is to estimate, for a given task, suitable hyperparameters within a flexible graph family.
In particular, we aim to infer 
\bit
\item $\bA$, containing the bandwidths (or weights) $a_m$'s for different dimensions in Eq. \eqref{wrbf}, as well as 
\item $k$, for sparse $k$NN graph construction;
\eit
so as to 
better align the graph structure with the underlying (hidden) data manifold and the given SSL task.

%% file: 03proposed.tex
In this section, we present the formulation and efficient computation of our graph learning algorithm \method, for Parallel Graph Learning for SSL.

In essence, the feature weights $a_m$'s and $k$ are the model parameters that govern how the algorithm's performance generalizes to unlabeled data. Typical model selection approaches include random search or grid search to find a configuration of the hyperparameters that yield the best cross-validation performance.

Unfortunately, the search space becomes prohibitively large for high-dimensional datasets that could render such methods futile.
In such cases, one could instead carefully select the configurations in an adaptive manner.
The general idea is to impose a smooth loss function $g(\cdot)$ on the validation set over which $\bA$ can be estimated using a gradient based method.

We present the
main steps of our algorithm for adaptive hyper-parameter search in Algorithm \ref{alg:meta}.

\begin{algorithm}
	\caption{{\sc Gradient} (for Adaptive Hyperparameter Search)\label{alg:meta}}
	\begin{algorithmic}[1]
		\STATE Initialize $k$ and $\ba$ (vector containing $a_{m}$'s); $\;\;t:=0$
		\REPEAT
		\STATE Compute $\bF^{(t)}$ using $k$NN graph on current $a_{m}$'s by \eqref{power}
		\STATE Compute gradient $\frac{\partial g}{\partial a_{m}}$ based on  $\bF^{(t)}$ by \eqref{gradg} for each $a_m$
		\STATE Update $a_{m}$'s by $\ba^{(t+1)} := \ba^{(t)} - \gamma \frac{\mathrm{d} g}{\mathrm{d} \ba}$;  $\;\;t:=t+1$
		\UNTIL  $a_{m}$'s have converged
	\end{algorithmic}
\end{algorithm}
\setlength{\textfloatsep}{0.0in}

The initialization in step 1 can be done using some heuristics, although the most prevalent and easiest approach is a random guess.
Given a fixed initial (random) configuration, we essentially perform an adaptive search that strives to find a better configuration around it, guided by the validation loss $g(\cdot)$. In Section \ref{ssec:gloss},
we introduce the specific function $g(\cdot)$ that we use and how to compute its gradient.

While the gradient based optimization is likely to find a better configuration than where it started, the final performance of the SSL algorithm depends considerably on the initialization.
Provided that the search space is quite large for high dimensional datasets, it is of paramount importance to try 
different random initializations in step 1, in other words, to run Algorithm \ref{alg:meta} several times.
As such, \textit{the \grad\ algorithm can be seen as an adaptive local search}, where we start at a random configuration and adaptively search in the vicinity for a better one.

As we discuss in Section \ref{ssec:gloss},  the gradient based updates are computationally demanding. This makes na\"ively running Algorithm \ref{alg:meta} several times expensive.
There are however two properties that we can take considerable advantage of: (1) both the SSL algorithm (using the power method) as well as the gradient optimization are \textit{iterative}, {\em any-time} algorithms (i.e., they can return an answer at any time that they are probed), and (2) different initializations can be run independently in \textit{parallel}.

In particular, our search strategy is inspired by a general framework of parallel hyperparameter search designed for \textit{iterative} machine learning algorithms that has been recently proposed by Jamieson and Talwalkar \cite{DBLP:conf/aistats/JamiesonT16} and a follow-up by Li et al. \cite{DBLP:journals/corr/LiJDRT16}. This framework perfectly suits our SSL setting for the reasons (1) and (2) above.
The idea is to start multiple (random) configurations in parallel threads, run them for a bounded amount of time, probe for their solutions, throw out the worst half (or some other pre-specified fraction), and repeat until one configurations remains.
By this strategy of early termination, that is by quitting poor initializations early without running them to completion, the compute resources are effectively allocated to  
promising hyperparameter configurations.
Beyond what has been proposed in \cite{DBLP:conf/aistats/JamiesonT16}, we start new initializations on the idle threads whose jobs have been terminated in order to fully utilize the parallel threads. 
We describe the details of our parallel search in Section \ref{ssec:parallel}.


%% file: 03validation.tex
\vspace{-0.1in}
\subsection{Validation Loss $g(\cdot)$ \& Gradient Updates}
\label{ssec:gloss}

We base the learning of the hyperparameters of our kernel function ($a_m$'s in Eq. \eqref{wrbf}) on minimizing some loss criterion on validation data.
Let $\mL \subset \mD$ denote the set of $l$ labeled examples, and $\mV \subset \mL$ a subset of the labeled examples designated as validation samples.
A simple choice for the validation loss would be the labeling error, written as
$g_{\bA}(\mV) =  \sum_{v\in \mV} (1-\bF_{vc_v})$, where $c_v$ denotes the true class index for a validation instance $v$.
Other possible choices for each $v$ include $-\log \bF_{vc_v}$, $(1-\bF_{vc_v})^x$, $x^{-\bF_{vc_v}}$, with $x>1$. 

In semi-supervised learning the labeled set is often small. This means the number of validation examples is also limited.
To squeeze the most out of the validation set, we propose to use a \textit{pairwise} learning-to-rank objective:
\begin{equation}
\label{rankobj}
g_{\bA}(\mV) = \sum_{c'=1}^c \;\; \sum_{(v,v'): \; v\in \mV_{c'}, \atop{v'\in \mV\backslash \mV_{c'}}} -\log \sigma(\bF_{vc'}-\bF_{v'c'})
\end{equation}
where $\mV_{c'}$ denotes the validation nodes whose true class index is $c'$ and $\sigma(x)=\frac{\exp(x)}{1+\exp(x)}$ is the sigmoid function.
The larger the difference $(\bF_{vc'}-\bF_{v'c'})$, or intuitively the more \textit{confidently} the solution $\bF$ ranks
validation examples of class $c'$ above other validation examples not in class $c'$, the better it is; since then $\sigma(\cdot)$ would approach 1 and the loss to zero.

In short, we aim to find the hyperparameters $\bA$ that minimize the total negative
log likelihood of ordered validation pairs. 
The optimization is conducted by gradient descent. 
The gradient is computed as
\begin{align}
	\label{gradg}
\frac{\partial g}{\partial a_{m}} & = \frac{\partial \bigg( \sum_{c'=1}^c\sum\limits_{(v,v'):v\in \mV_{c'}, {v'\in \mV\backslash \mV_{c'}}} - \bF_{vv'} + \log (1+\exp{(\bF_{vv'})}) \bigg)}{\partial a_{m}} \nonumber \\  
& = \sum_{c'=1}^c \sum\limits_{(v,v'):v\in \mV_{c'}, {v'\in \mV\backslash \mV_{c'}}} (o_{vv'}-1) 
\big( \frac{\partial \bF_{vc'}}{\partial a_{m}} - \frac{\partial \bF_{v'c'}}{\partial a_{m}} \big) 
\end{align}
where we denote by $\bF_{vv'} = (\bF_{vc'}-\bF_{v'c'})$ and $o_{vv'} = \sigma(\bF_{vv'})$.

The values 
$\frac{\partial \bF_{vc'}}{\partial a_{m}}$ and
$\frac{\partial \bF_{v'c'}}{\partial a_{m}}$ for each class $c'$ and $v,v'\in \mV$
can be read off of matrix $\frac{\partial \bF}{\partial a_{m}}$, which is given as
\begin{equation}
\label{eq:F}
\frac{\partial \bF}{\partial a_{m}} = - (\bI_n + \alpha \bL)^{-1} 
\frac{\partial (\bI_n + \alpha \bL)}{\partial a_{m}} \bF
= \alpha (\bI_n + \alpha \bL)^{-1}  \frac{\partial \bP}{\partial a_{m}} \bF\;,
\end{equation}
using the equivalence $d\bX^{-1} = -\bX^{-1}(d\bX) \bX^{-1}$.
Recall that $\bP=\bD^{-1/2}\bW\bD^{-1/2}$ with $\bP_{ij} = \frac{\bW_{ij}}{\sqrt{d_id_j}}$; $d_i$ being node $i$'s degree  in $\mG$.

We can then write
\begin{align}
	\frac{\partial \bP_{ij}}{\partial a_{m}} =& \frac{\partial \bW_{ij}}{\partial a_{m}}\frac{1}{\sqrt{d_id_j}} - \frac{\bW_{ij}}{2}(d_id_j)^{-3/2}\frac{\partial d_id_j}{\partial a_{m}} \\
=& \frac{\partial \bW_{ij}}{\partial a_{m}}\frac{\bP_{ij}}{\bW_{ij}} - \frac{\bW_{ij}}{2}(\frac{\bP_{ij}}{\bW_{ij}})^{3}(d_j\frac{\partial d_i}{\partial a_{m}}+d_i\frac{\partial d_j}{\partial a_{m}}) \\
=& \frac{\partial \bW_{ij}}{\partial a_{m}}\frac{\bP_{ij}}{\bW_{ij}} - \frac{\bW_{ij}}{2}(\frac{\bP_{ij}}{\bW_{ij}})^{3} \nonumber\\
& \big(\sum_n\bW_{in}\cdot \sum_n\frac{\partial \bW_{jn}}{\partial a_{m}} + \sum_n\bW_{jn}\cdot \sum_n\frac{\partial \bW_{in}}{\partial a_{m}} \big)	\label{amder}
\end{align}

\subsubsection{Matrix-form gradient}
We can rewrite all element-wise gradients into a combined matrix-form gradient. 
The matrix-form is compact and can be computed more efficiently on platforms optimized for matrix operations (e.g., Matlab).

The matrix-form uses 3-d matrix (or tensor) representation. In the following, we use $\odot$ to denote element-wise multiplication, $\oslash$ element-wise division, and $\otimes$ for element-wise power. In addition,  $\cdot$ denotes regular matrix dot product. For multiplication and division, a 3-d matrix should be viewed as a 2-d matrix with vector elements.  

First we extend the derivative w.r.t. $a_m$ in Eq. \eqref{amder} into derivative w.r.t. $\ba$:
\begin{align}
\label{ew}
	\frac{\partial \bP_{ij}}{\partial \ba} = & \; \frac{\partial \bW_{ij}}{\partial \ba}\frac{\bP_{ij}}{\bW_{ij}} - \frac{\bW_{ij}}{2}  (\frac{\bP_{ij}}{\bW_{ij}})^{3}
	\nonumber \\
	&(\sum_n\bW_{in}\cdot \sum_n\frac{\partial \bW_{jn}}{\partial \ba} + \sum_n\bW_{jn}\cdot \sum_n\frac{\partial \bW_{in}}{\partial \ba})
\end{align}

To write this equation concisely, let tensor $\Omega$ be $\frac{\partial \bW}{\partial \ba}$, a 2d-matrix with vector elements $\Omega_{ij} = \frac{\partial \bW_{ij}}{\partial \ba}$, and let tensor $\Delta \bX$ be the one with vector elements $\Delta \bX_{ij} = (\bx_i - \bx_j)^2$. 

Then we can rewrite some equations using the above notation:
\begin{align}
	\sum_n \bW_{in} & = (\bW \cdot \bone_n)_i \\
	 \sum_n \frac{\partial \bW_{jn}}{\partial \ba} &= (\Omega \cdot \bone_n)_j \\
	 \sum_n \bW_{in} \cdot \sum_n \frac{\partial \bW_{jn}}{\partial \ba} &=  (\bW\cdot \bone_n \cdot (\Omega \cdot \bone_n)^T )_{ij}
\end{align}
Now we can rewrite element-wise gradients in \eqref{ew} into one matrix-form gradient:
\begin{align}
	\label{tensor}
	\frac{\mathrm{d} \bP}{\mathrm{d} \ba} = & \; \Omega \odot (\bP \oslash \bW) - \frac{1}{2}\bP^{\otimes 3} \oslash \bW^{\otimes 2} \nonumber \\
	& \odot( \bW\cdot \textbf{1}_n \cdot (\Omega \cdot \textbf{1}_n)^T + (\bW\cdot \textbf{1}_n \cdot (\Omega \cdot \textbf{1}_n)^T )^T)
\end{align}

The only thing left is the computation of $\Omega = \frac{\partial \bW}{\partial \ba}$.
Notice that
\begin{align}
\frac{\partial \bW_{ij}}{\partial a_{m}} =&  \frac{\partial \exp(- \sum_{m=1}^d a_{m} (\bx_{id} - \bx_{jd})^2 ) }{\partial a_{m}}
= - \bW_{ij} (\bx_{id} - \bx_{jd})^2 \nonumber \\
\Longrightarrow & \;\; \frac{\partial \bW_{ij}}{\partial \ba}  = -\bW_{ij}(\bx_{i} - \bx_{j})^2 = -\bW_{ij}\Delta \bX_{ij} \nonumber \\
\Longrightarrow & \;\; \frac{\mathrm{d} \bW}{\mathrm{d} \ba} = -\bW\odot\Delta \bX = \Omega \label{omega}
\end{align}

All in all, we transform the element-wise gradients $\frac{\partial \bP_{ij}}{\partial a_{m}}$ as given in Eq. \eqref{amder} to compact tensor-form updates $\frac{\mathrm{d}\bP}{\mathrm{d}\ba}$ as in Eq. \eqref{tensor}. The tensor-form gradient updates not only provide speed up, but also can be expanded to make full use of the $k$NN graph sparsity. In particular, $\bW$ is a $k$NN-sparse matrix with $O(kn)$ non-zero elements. First, Eq. \eqref{omega} for $\Omega$ shows that we do not need to compute full $\Delta \bX$ but only the elements in $\Delta \bX$ corresponding to non-zero elements of $\bW$. Similarly, in Eq. \eqref{tensor}, matrix $\bP$ does not need to be fully computed, and the whole Eq. \eqref{tensor} can be computed sparsely. 



\subsubsection{Complexity analysis}
We first analyze computational complexity in terms of two main components: constructing the $k$NN graph and computing $\bF$ in line 3, and
computing the gradient $\frac{\mathrm{d} g}{\mathrm{d} \ba}$ in line 4 of Algorithm \ref{alg:meta} as outlined in this subsection.

Let us  denote the number of non-zeros in $\bW$, i.e. the number of edges in the $k$NN graph, by $e = nnz(\bW)$. We assume $kn \leq e \leq 2kn$ remains near-constant as $\ba$ changes over the \grad~ iterations.

In line 4, we first construct tensor $\Omega$ as in Eq. \eqref{omega} in $O(ed)$.
Computing $\frac{\mathrm{d} \bP}{\mathrm{d} \ba}$ as in Eq. \eqref{tensor} also takes $O(ed)$.
Next, obtaining matrix $\frac{\partial \bF}{\partial a_{m}}$ in Eq. \eqref{eq:F}
seemingly requires inverting $(\bI_n + \alpha \bL)^{-1}$. However, we
￼can rewrite Eq. \eqref{eq:F} as
$$
(\bI_n+\alpha \bI_n - \alpha \bP) \frac{\partial \bF}{\partial a_{m}} = \alpha \frac{\partial \bP}{\partial a_{m}} \bF \Rightarrow
\frac{\partial \bF}{\partial a_{m}} = \alpha (\bP-\bI_n) \frac{\partial \bF}{\partial a_{m}} + \alpha \frac{\partial \bP}{\partial a_{m}} \bF
$$
which can be solved via the power method that takes $t$ iterations in $O(ect)$.
Computing $\frac{\partial \bF}{\partial a_{m}}$ and plugging in Eq. \eqref{gradg} to get $g(\cdot)$'s gradient for all $a_m$'s then takes $O(ectd)$, or equivalently $O(knctd)$.


In line 3, updated $a_m$'s are used for weighted node similarities to compute $k$NNs for each instance. 
Nearest neighbor computation for all instances is inherently quadratic, which however can be sped up by approximation algorithms and data structures such as locality-sensitive hashing (LSH) \cite{lsh}. To this end, we use a fast $k$NN graph construction algorithm that takes advantage of LSH and has $O(n[dk^2+\log n])$ complexity \cite{conf/pkdd/ZhangHGL13}; only quadratic in the (small) $k$ but log-linear in $n$. 
Given the $k$NN graph, $\bF$ can then be computed via \eqref{power} in $O(ect')$ for $t'$ iterations of the power method.

Overall, one iteration of Algo. \ref{alg:meta} takes $O(n[kctd + dk^2+\log n])$. Furthermore, if we consider $k,c,t$ as constants, then the computational complexity can be written as $O(n[d+\log n])$.

In addition, memory requirement for each gradient update is $O(knd)$. The bottleneck is the construction of tensors $\Omega$ and $\Delta\bX$ with size-$d$ vector elements. As discussed earlier those are constructed sparsely, i.e., only the elements corresponding to non-zero entries of $\bW$, which is $O(kn)$, are stored.

%% file: 03parallel.tex
\subsection{Parallel Hyperparameter Search with Adaptive Resource Allocation}
\label{ssec:parallel}

For high-dimensional datasets, the search space of hyperparameter configurations is huge.
In essence, Algorithm \ref{alg:meta} is an adaptive search around a single initial point in this space.
As with many gradient-based optimization of high-dimensional non-convex functions with unknown smoothness,
its performance depends on the initialization.
Therefore, trying different  initializations of Algorithm \ref{alg:meta} is beneficial to improving performance.

\begin{figure}[!t]
	\begin{tabular}{c} 
		\includegraphics[width=0.45\textwidth]{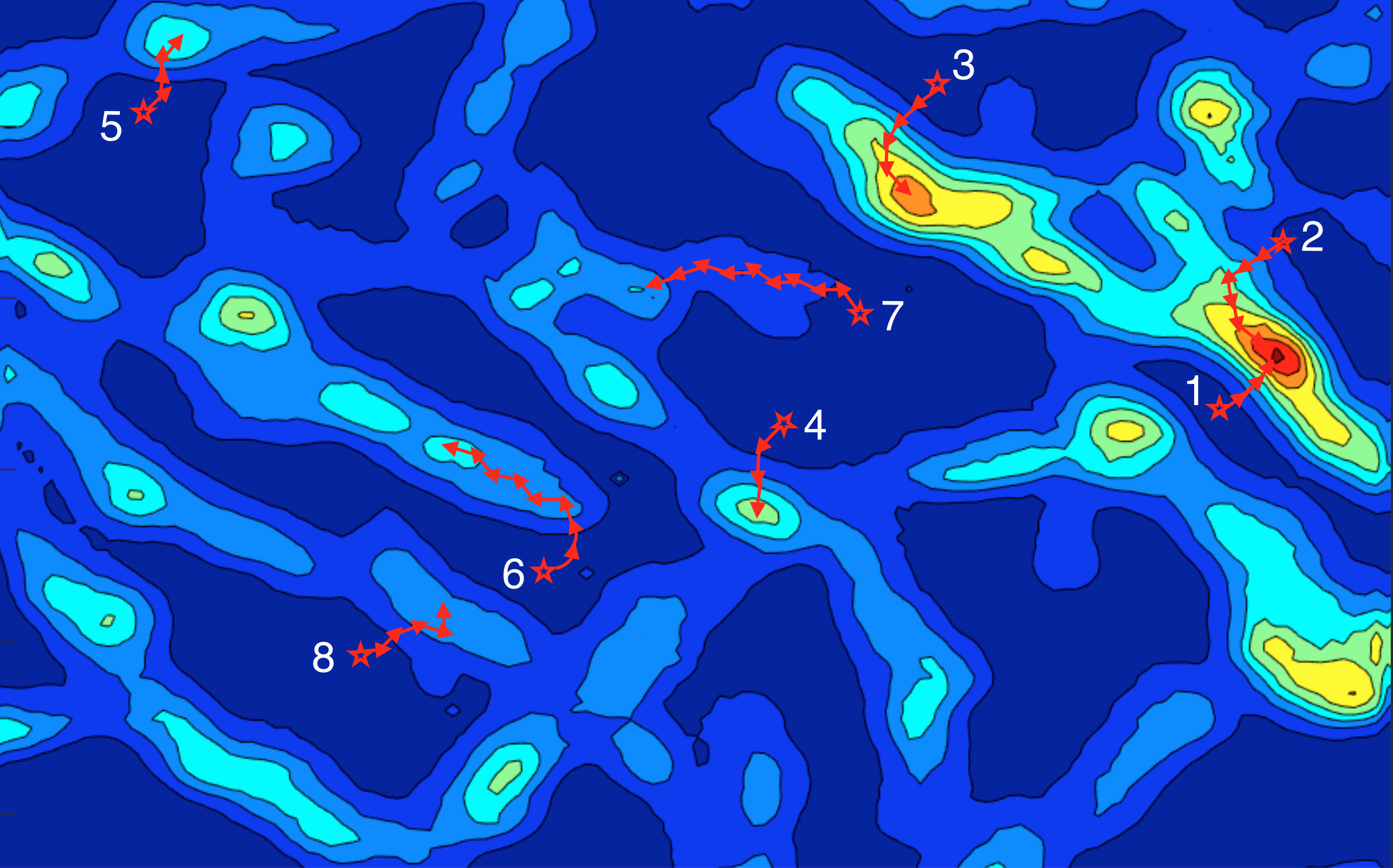} 
	\end{tabular}
		 \vspace{-0.1in}
	\caption{(best in color) The heatmap shows the validation error over an example 2-d search space with red corresponding to areas with lower error. Our approach is an inter-mix of random and adaptive search. We start at various random configurations (stars 1--8) and adaptively improve them (arrows depicting gradient updates), while strategically terminating unpromising ones (like 6, 7, and 8) early.}
	\label{fig:randadap}
	  \vspace{0.1in}
\end{figure}

An illustrative example over a 2-d search space is shown in Figure \ref{fig:randadap} (best in color).
In this space most configurations yield poor validation accuracy, as would be the likely case in even higher dimensions.
In the figure, eight random configurations are shown (with stars). The sequence of arrows from a configuration can be seen analogous to the iterations of a single run of Algorithm \ref{alg:meta}.

While it is beneficial to try as many random initializations as possible, especially in high dimensions,
adaptive search is slow. A single gradient update by Algorithm \ref{alg:meta} takes time in $O(ectd)$ followed by reconstruction of the $k$NN graph.
Therefore, it would be good to quit the algorithm early if the ongoing progress is not promising (e.g., poor initializations 6--8 in Figure \ref{fig:randadap}) and simply try a new initialization.
This would allow using the time efficiently for going through a larger number of configurations.

One way to realize such a scheme is called successive halving \cite{DBLP:conf/aistats/JamiesonT16}, which relies on an early-stopping strategy for iterative machine learning algorithms.
The idea is quite simple and follows directly from its name: try out a set of hyperparameter configurations for some fixed amount of time (say in parallel threads), evaluate the performance of all configurations, keep the best half (terminate the worst half of the threads), and repeat until one configuration remains while allocating exponentially increasing amount of time after each round to not-yet-terminated,  promising configurations (i.e., threads).
Our proposed method is a parallel implementation of their general framework adapted to our problem, and further utilizes the idle threads that have been terminated.

Algorithm \ref{alg:pglearn} gives the steps of our proposed method \method, which calls the {\sc Gradient} subroutine in Algorithm \ref{alg:meta}.
Besides the input dataset $\mD$, \method~requires three inputs: (1) budget $B$; the maximum number of time units\footnote{We assume time is represented in units, where a unit is the minimum amount of time one would run the models before comparing against each other.} that can be allocated to one thread (i.e., one initial hyperparameter configuration),
(2) downsampling rate $r$; an integer that controls the fraction of threads terminated (or equally, configurations discarded) in each round of \method, and finally (3) $T$; the number of parallel threads.

Concretely, \method~performs $R=\lfloor \log_r B \rfloor$ rounds of elimination. At each round, the best $1/r$ fraction of configurations are retained. Eliminations are done in exponentially increasing time intervals, that is, first round occurs at time $B/r^R$, second round  at $B/r^{R-1}$, and so on. 

After setting the number of elimination rounds $R$ and the duration of the first round, denoted $d_1$ (line 1),
\method~starts by obtaining $T$ initial hyperparameter configurations (line 2). Note that a configuration is a $(k,\ba_{1:d})$ pair. Our implementation\footnote{\label{code}We release all source code at {\url{https://github.com/LingxiaoShawn/PG-Learn}}} 
of \method~is parallel. As such, each thread draws their own configuration; uniformly at random.
Then, each thread runs the \grad~(Algorithm \ref{alg:meta}) with their corresponding configuration for duration $d_1$ and returns the validation loss (line 3).

At that point, \method~enters the rounds of elimination (line 4).
$L$ validation loss values across threads are gathered at the master node,
which identifies the top $\lfloor T / r \rfloor$ configurations $\mC_{top}$  (or threads) with the lowest loss (line 5).
The master then terminates the runs on the remaining threads and restarts them afresh with new configurations $\mC_{new}$ (line 6). The second round is to run until $B/r^{R-1}$, or for $B/r^{R-1} - B/r^{R}$ in duration.
After the $i$th elimination, in general, we run the threads for duration $d_i$ as given in line 7---notice that exponentially increasing amount of time is provided to ``surviving'' configurations over time.
In particular, the threads with the promising configurations in $\mC_{top}$ are {\em resumed} their runs from where they are left off with the \grad~iterations (line 8). The remaining threads start with the \grad~iterations using their new initialization (line 9). Together, this ensures full utilization of the threads at all times.
Eliminations continue for $R$ rounds, following the same procedure of resuming best threads and restarting from the rest of the threads (lines 4--11).
After round $R$, all threads run until time budget $B$ at which point the (single) configuration with the lowest validation error is returned (line 12).

\begin{algorithm}[!t]
	\caption{\method~(for Parallel Hyperparameter Search)}
	\label{alg:pglearn}
	\begin{algorithmic}[1]
		\REQUIRE Dataset $\mD$, budget $B$ time units, downsampling rate $r$ ($=2$ by default), number of parallel threads $T$
		\ENSURE Hyperparameter configuration $(k,\ba)$ 
		\STATE  $R = \lfloor \log_r B \rfloor$, $d_1 = B r^{-R}$
		\STATE  $\mC := $ {\tt get\_hyperparameter\_configuration}$(T)$ 
		\STATE $L:= $ $\{${\tt run\_}{\sc Gradient\_}{\tt then\_return\_val\_loss}$(c,d_1) : c\in \mC\}$
		\FOR{$i\in \{1,\ldots, R\}$}
		\STATE $\mC_{top} := $ {\tt get\_top}$(\mC, L, \lfloor T / r \rfloor)$
		\STATE $\mC_{new} :=$ {\tt get\_hyperparameter\_configuration}$(T-\lfloor T/r \rfloor)$ \\
		\STATE $d_i = B (r^{-(R-i)} - r^{-(R-i+1)})$
		\STATE $L_{top}:=\{${\tt resume\_}{\sc Gradient\_}{\tt then\_return\_val\_loss}$(c,d_i)$ for $c\in \mC_{top}\}$
		\STATE $L_{new}:=\{${\tt run\_}{\sc Gradient\_}{\tt then\_return\_val\_loss}$(c,d_i)$ for $c\in \mC_{new}\}$
		\STATE $\mC := \mC_{top} \cup \mC_{new}, \;\; L:= L_{top} \cup L_{new}$
		\ENDFOR
		\RETURN $c_{top} := $ {\tt get\_top}$(\mC, L, 1)$
	\end{algorithmic}
\end{algorithm}
\setlength{\textfloatsep}{0.25in}

The underlying principle of \method~exploits the insight that a hyperparameter configuration which is destined to yield good performance ultimately is more likely than not to perform in the top fraction of configurations even after a small number of iterations. In essence, even if the performance of a configuration after a small number of iterations of the \grad~(Algorithm 1) may not be representative of its ultimate performance after a large number of iterations in {\em absolute} terms, its \textit{relative} performance in comparison to the alternatives is roughly maintained. We note that different configurations get to run different amounts of time before being tested against others for the first time (depending on the round they get introduced). This diversity  offers some robustness against variable convergence rates of the $g(\cdot)$ function at different random starting points.

\begin{figure}[h]
	\begin{tabular}{c} 
		\includegraphics[width=0.45\textwidth]{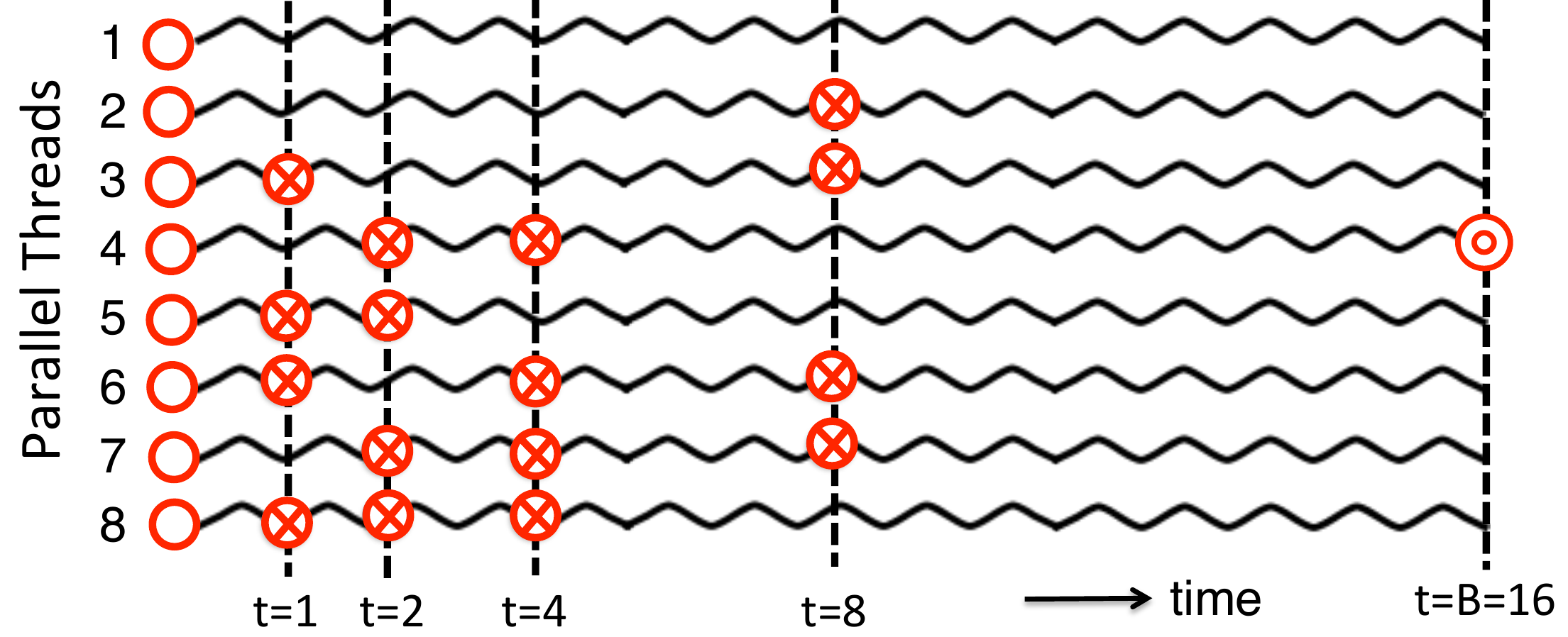} 
	\end{tabular}
		\vspace{-0.1in}
	\caption{Example execution of \method~ with $T=8$ parallel threads, downsampling rate $r=2$, and budget $B=16$ time units.
	At each ``check point'' in time (dashed vertical lines), (worst) half of the runs are discarded and corresponding threads restart Algorithm \ref{alg:meta} with new random configurations of ($k$, $\ba_{1:d})$. At the end, hyperparameters that yield the lowest $g(\cdot)$ function value (i.e. validation loss) across all threads are returned (those by thread 4 in this example).}
	\label{fig:demo}
\end{figure}

{\bf Example:}
In Figure \ref{fig:demo} we provide a simple example to illustrate \method's execution, using
$T=8$ parallel threads, downsampling rate $r=2$ (equiv. to halving), and $B=16$ time units of processing budget.
There are $\lfloor \log_2 16 \rfloor = 4$ rounds of elimination at $t=1,2,4,8$ respectively, with the final selection being made at $t=B$.
It starts with 8 different initial configurations (depicted with circles) in parallel threads.
At each round, bottom half (=4) of the threads with highest validation loss are terminated with their iterations of Algorithm 1 and restart running Algorithm 1 with a new initialization (depicted with a crossed-circle).
Overall, $T + (1-1/r) T \lfloor \log_r B \rfloor =8+4\lfloor \log_2 16 \rfloor = 24$ configurations are examined---a larger number as compared to the initial 8, thanks to the early-stopping and adaptive resource allocation strategy.

\begin{figure}[t]
	\begin{tabular}{cc} 
		\hspace{-0.2in}	\includegraphics[width=0.25\textwidth,height=1.3in]{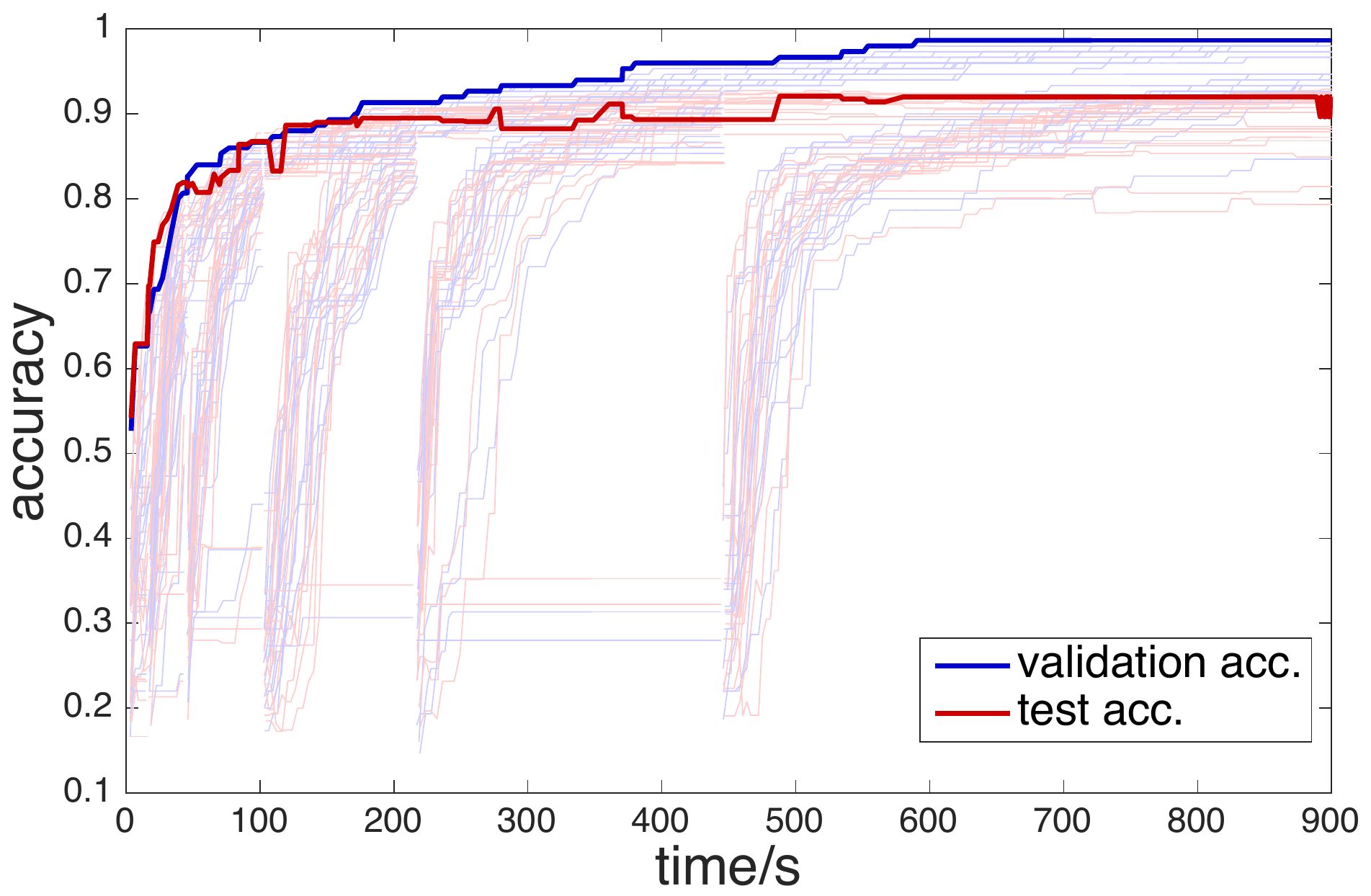} &
		\hspace{-0.2in}	\includegraphics[width=0.25\textwidth,height=1.3in]{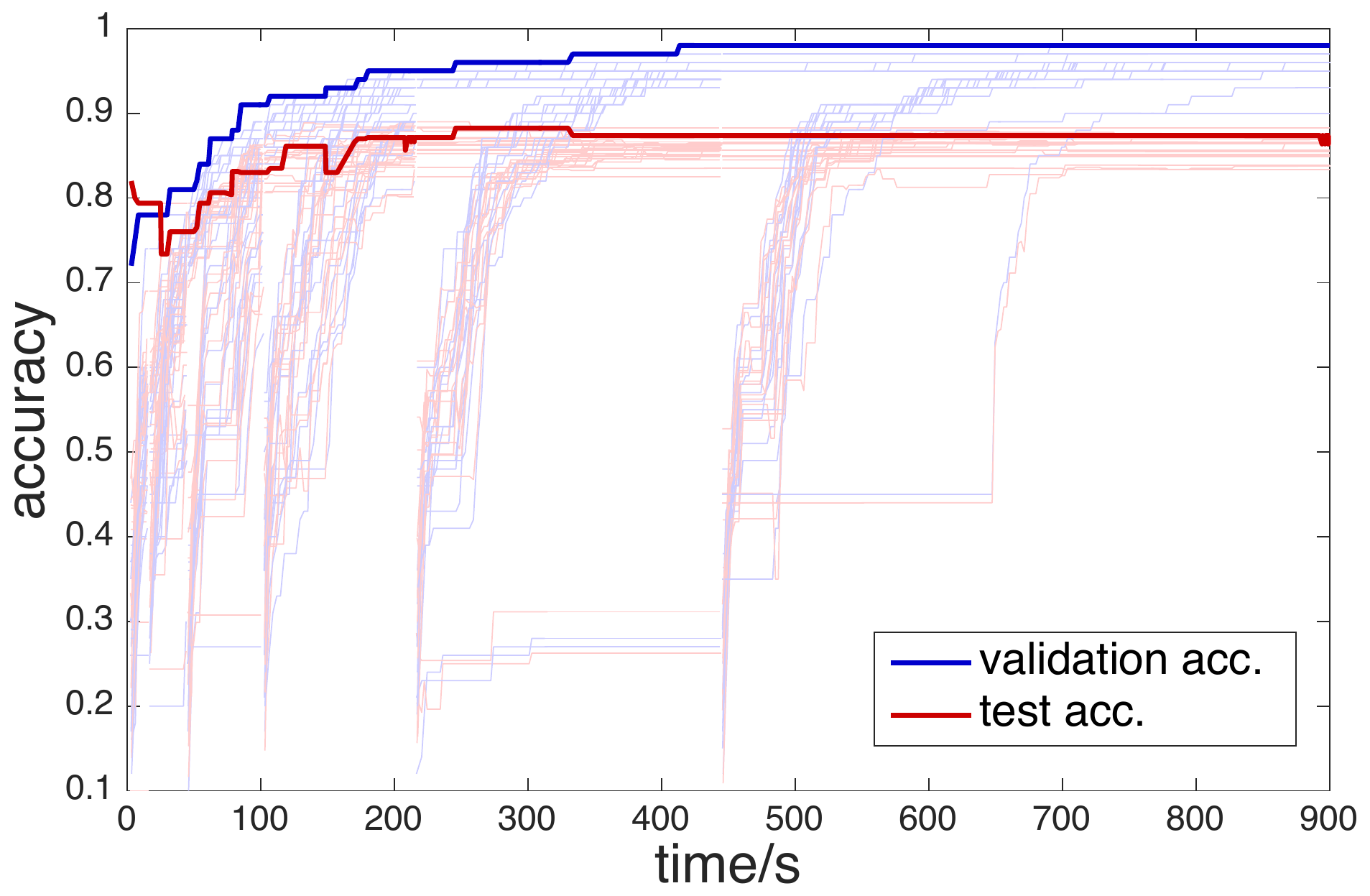} 
	\end{tabular}
	 \vspace{-0.1in}
	\caption{\method's val. (blue) and corresponding test (red) acc. vs. time on \coil~(left) and \mnist~(right) (see Table \ref{tab:data}).}
	\label{fig:exrun}
\end{figure}

Next in Figure \ref{fig:exrun} we show example runs on two different real-world datasets, depicting the progression of validation (blue) and test (red) accuracy over time, using $T=32, r=2, B=64$; $\approx$15 sec. unit-time. Thin curves depict those for individual threads. Notice the new initializations starting at different rounds, which progressively improve their validation accuracy over gradient updates (test acc. closely follows). Bold curves depict the overall-best validation accuracy (and corresponding test acc.) across all threads over time.

{\bf Setting $T$, $B$, and $r$:} Before we conclude the description of our proposed method, we briefly discuss the choices for its inputs. 
Number of threads $T$ is a resource-driven input. Depending on the platform being utilized---single machine or a parallel architecture like Hadoop or Spark---\method~can be executed with as many parallel threads as physically available to the practitioner.
Time units $B$ should be chosen based on the upper bound of practically available time.
For example, if one has time to run hyperparameter tuning for at most 3 hours and the minimum amount of time that is meaningful to execute gradient search of configurations before comparing them (i.e., unit time) is 5 minutes, then $B$ becomes $180/5 = 36$ units.
Finally, $r$ can be seen as a knob for greediness. A larger value of $r$ corresponds to more aggressive elimination with fewer rounds; specifically, each round terminates $T(r-1)/r$ configurations for a total of $\lfloor \log_r B \rfloor$ rounds.
All in all, $T$ and $B$ are set based on practical {\em resource constraints}, physical and temporal, respectively. On the other hand, $r$ can be set to a small integer, like $2$ or $3$, without results being very sensitive to the choice.

%% file: 03experiments.tex
\subsection{Datasets and Baselines}
\label{ssec:databaselines}

{\bf Datasets:} We use the publicly available multi-class classification datasets listed in Table \ref{tab:data}. \coil\footnote{{\footnotesize{\url{http://olivier.chapelle.cc/ssl-book/index.html}}, see `benchmark datasets'}} (Columbia Object Image Library) contains images of different objects taken from various angles. Features are shuffled and downsampled pixel values from the red channel.
\usps\footnote{\footnotesize{\url{http://www.cs.huji.ac.il/~shais/datasets/ClassificationDatasets.html}}}
is a standard dataset for handwritten digit recognition, with numeric pixel values scanned from the handwritten digits on envelopes from the U.S. Postal Service. 
\mnist\footnote{\footnotesize{\url{http://yann.lecun.com/exdb/mnist/}}} 
is another popular handwritten digit dataset, containing size-normalized and centered digit images. 
\umist\footnote{\footnotesize{\url{https://www.sheffield.ac.uk/eee/research/iel/research/face}}}  
face database is made up of images of 20 individuals with mixed race, gender, and appearance. Each individual takes a range of poses, from profile to frontal views.
\yale\footnote{\label{note1}\footnotesize{\url{http://www.cad.zju.edu.cn/home/dengcai/Data/FaceData.html}}} 
is a subset of the extended Yale Face Database B, which consists of frontal images under different illuminations from 5 individuals.

{\bf Baselines:} 
We compare the accuracy of \method~against five baselines that use a variety of schemes,
including the strawmen grid search and random guessing strategies, the seminal unsupervised gradient-based graph learning by Zhu et al., a self-representation based graph construction, and a metric learning based scheme.
Specifically,
\bit
\item[(1)] \grid~search (GS): $k$-NN graph with RBF kernel where $k$ and bandwidth $\sigma$ are chosen via grid search,
\item[(2)] \randd~search (RS): $k$-NN with RBF kernel where $k$ and different bandwidths $\ba_{1:d}$ are randomly chosen,
\item[(3)] \minent: Minimum Entropy based tuning of $\ba_{1:d}$'s as proposed by Zhu et al. \cite{zhuGhahramLaff03semisup} (generalized to multi-class), 
\item[(4)] 
\aew: Adaptive Edge Weighting by Karasuyama et al. \cite{journals/ml/KarasuyamaM17} that estimates $\ba_{1:d}$'s through local linear reconstruction, and
\item[(5)] \metric: Iterative self-learning scheme combined with distance metric learning by Dhillon et al. \cite{dhillon_acl10}.
\eit

Note that \grid\ and \randd\ are standard techniques employed by practitioners most typically.
\minent\ is perhaps the first graph-learning strategy for SSL which was proposed as part of the Gaussian Random Fields SSL algorithm.
It estimates hyperparameters by minimizing the entropy of the solution on unlabeled instances via gradient updates.
\metric\ uses and iteratively enlarges the labeled data (via self-learning) to estimate the metric $\bA$; which we restrict to a diagonal matrix, as our datasets are high dimensional and metric learning is prohibitively expensive for a full matrix. We generalized these baselines to multi-class and implemented them ourselves. We open-source (1)--(4) along with our \method~implementation.\footref{code}
Finally, \aew\ is one of the most recent techniques on graph learning, which extends the LLE \cite{roweis2000ndr} idea  by restricting the regression coefficients (i.e., edge weights) to be derived from Gaussian kernels.  We use their publicly-available implementation.\footnote{\label{note2}\footnotesize{\url{http://www.bic.kyoto-u.ac.jp/pathway/krsym/software/MSALP/MSALP.zip}}}

\begin{table}[!t]
	\centering
	\caption{Summary of (multi-class) datasets used in this work. }
	\label{tab:data}
	\vspace{-0.15in}
	\small{
		\begin{tabular}{l|rrrr} 
			\toprule
			\textbf{Name}   & \textbf{\#pts $n$} & {\bf \#dim $d$} & {\bf \#cls $c$}  & {\bf description}	\\ 
			\midrule		
			\coil & 1500 & 241 & 6 & objects with various shapes  \\
			\usps & 1000 & 256 & 10 & handwritten digits\\
			\mnist & 1000 & 784 & 10 & handwritten digits \\
			\umist & 575 & 644 & 20 & faces (diff. race/gender/etc.) \\
			\yale & 320 & 1024 & 5 & faces (diff. illuminations)\\
			\bottomrule		
		\end{tabular}}
		 \vspace{-0.1in}
	\end{table}

\subsection{Empirical Results}

\subsubsection{Single-thread Experiments} We first evaluate the proposed \method~ against the baselines on a fair ground  using a single thread, since the baselines do not leverage any parallelism. Single-thread \method~is simply the \grad~as given in Algo. \ref{alg:meta}.
	
{\bf Setup:} For each dataset, we sample 10\% of the points at random as the labeled set $\mL$, under the constraint that all classes must be present in $\mL$
and treat the remaining unlabeled data as the test set.
For each dataset, 10 versions with randomly drawn labeled sets are created and the average test accuracy across 10 runs is reported.
Each run starts with a different random configuration of hyperparameters.
For \method, \grid, and \randd, we choose (a small) $k \in [5,20]$. 
$\sigma$ for \grid~ and \minent\footnote{\minent~ initializes $\ba$ uniformly, i.e., all $a_m$'s are set to the same $\sigma$ initially \cite{zhuGhahramLaff03semisup}.}, and $a_m$'s for \method, \randd, and \aew~ are chosen from $[0.1\bar{d}, 10\bar{d}]$, where $\bar{d}$ is the mean Euclidean distance across all pairs.
Other hyperparameters of the baselines, like $\epsilon$ for \minent~and $\gamma$ and $\rho$ for \metric, are chosen as in their respective papers.
Graph learning is performed for 15 minutes, around which all gradient-based methods have converged.

{\bf Results:} Table \ref{tab:seq} gives the average test accuracy of the methods on each dataset, avg'ed over 10 runs with random labeled sets. \method~outperforms its competition significantly, according to the paired Wilcoxon signed rank test on a vast majority of the cases---only on the two handwritten digit recognition tasks there is no significant difference between \method~ and \minent.
Not only \method~ is significantly superior to existing methods, its performance is desirably high in absolute terms. 
It achieves 93\% prediction accuracy on the 20-class \umist, and 82\% on the $2^{10}$-dimensional \yale~ dataset.

\begin{figure}[!t]
	\centering
	\begin{tabular}{ccc}
		\hspace{-.1in}\includegraphics[width=0.35\columnwidth]{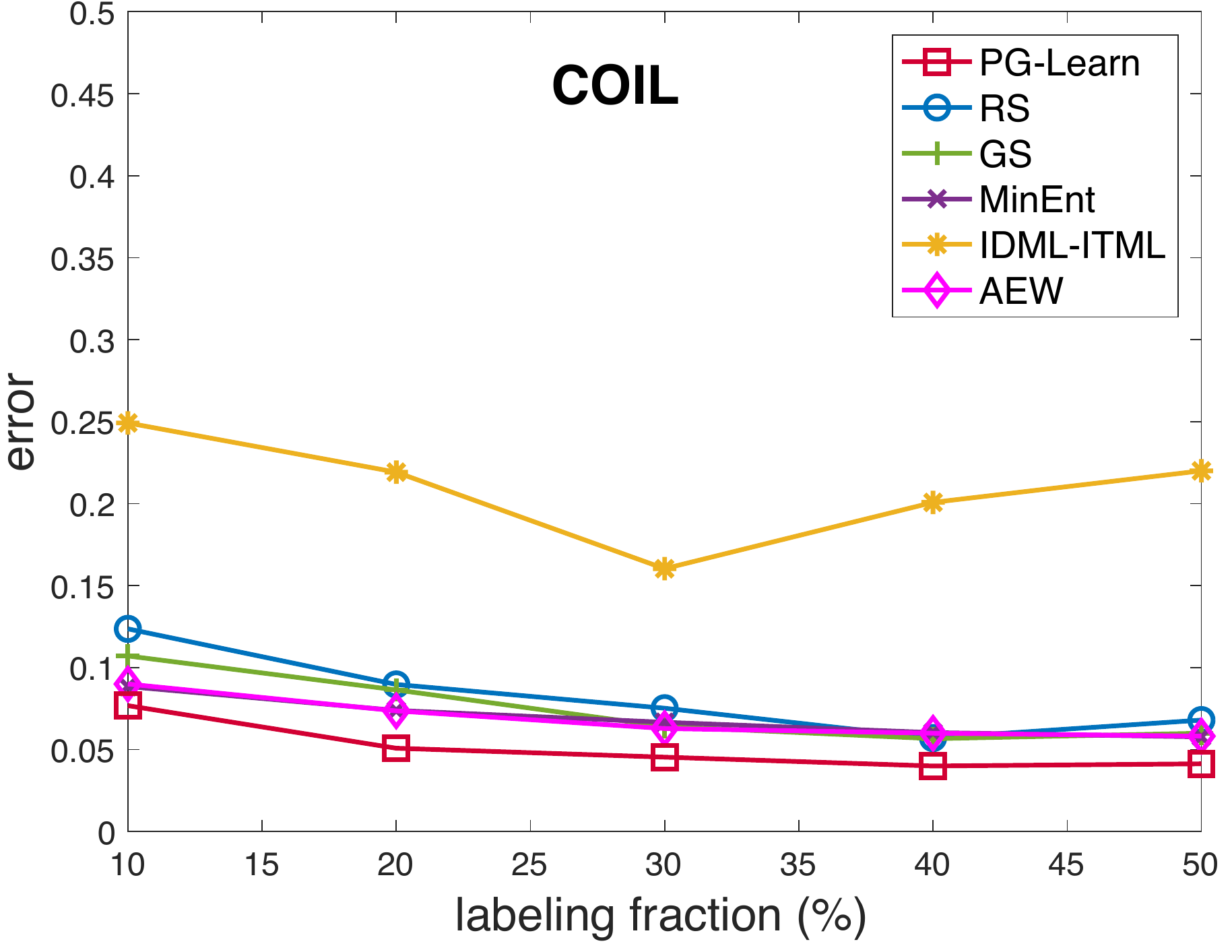}&
		\hspace{-.135in}\includegraphics[width=0.35\columnwidth]{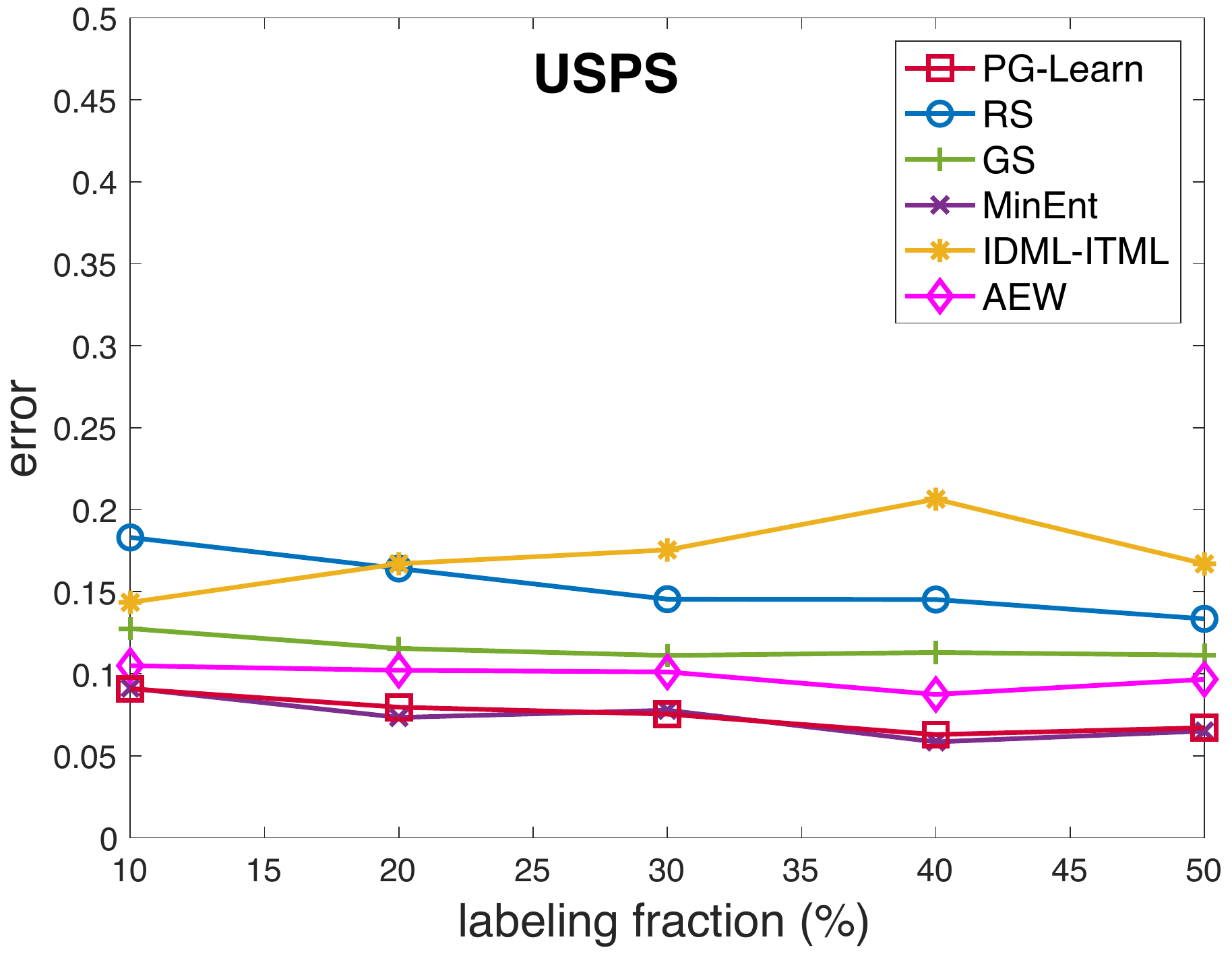}&
		\hspace{-.135in}\includegraphics[width=0.35\columnwidth]{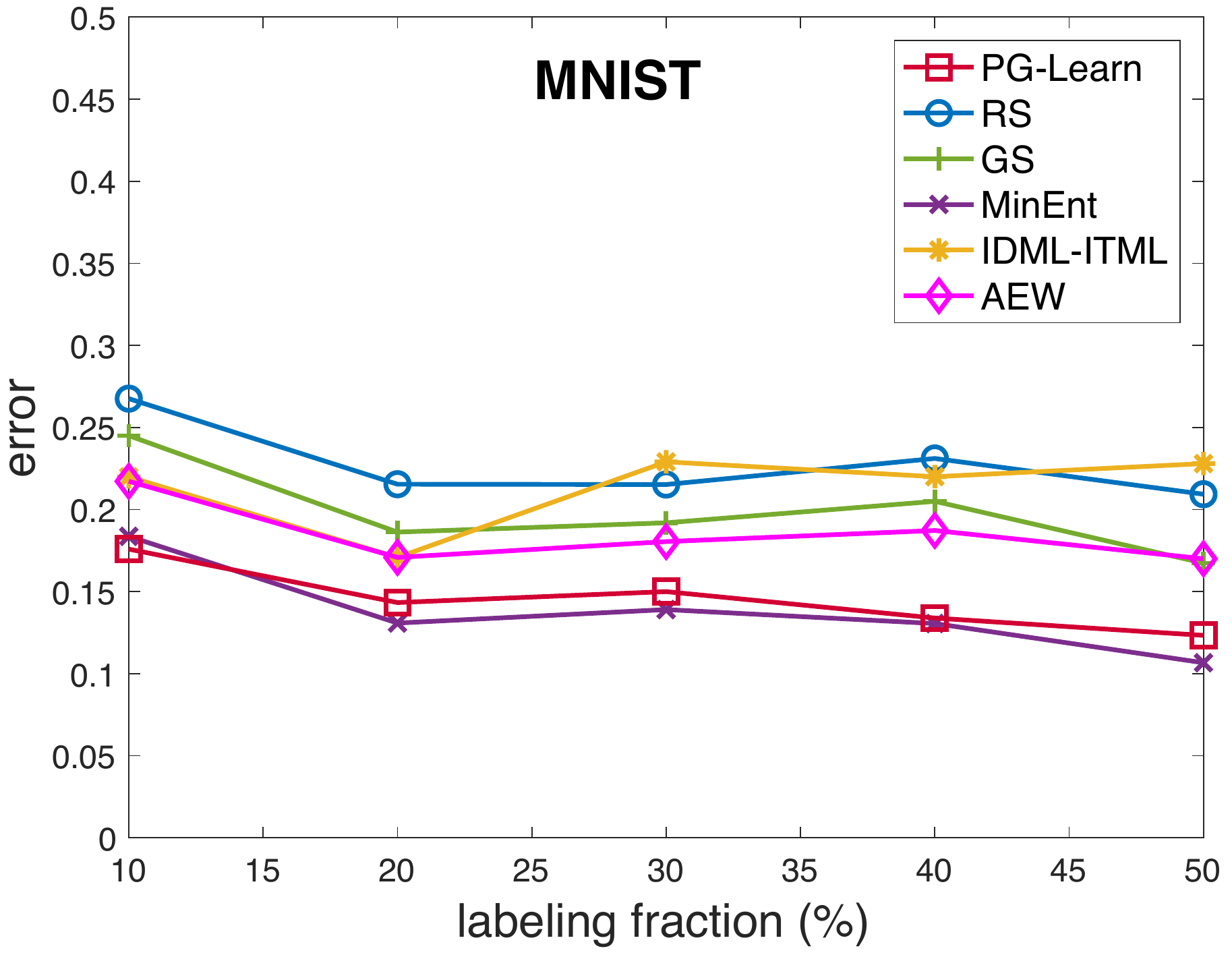}\\
	\end{tabular}
	\begin{tabular}{cc}
		\hspace{-.05in}\includegraphics[width=0.35\columnwidth]{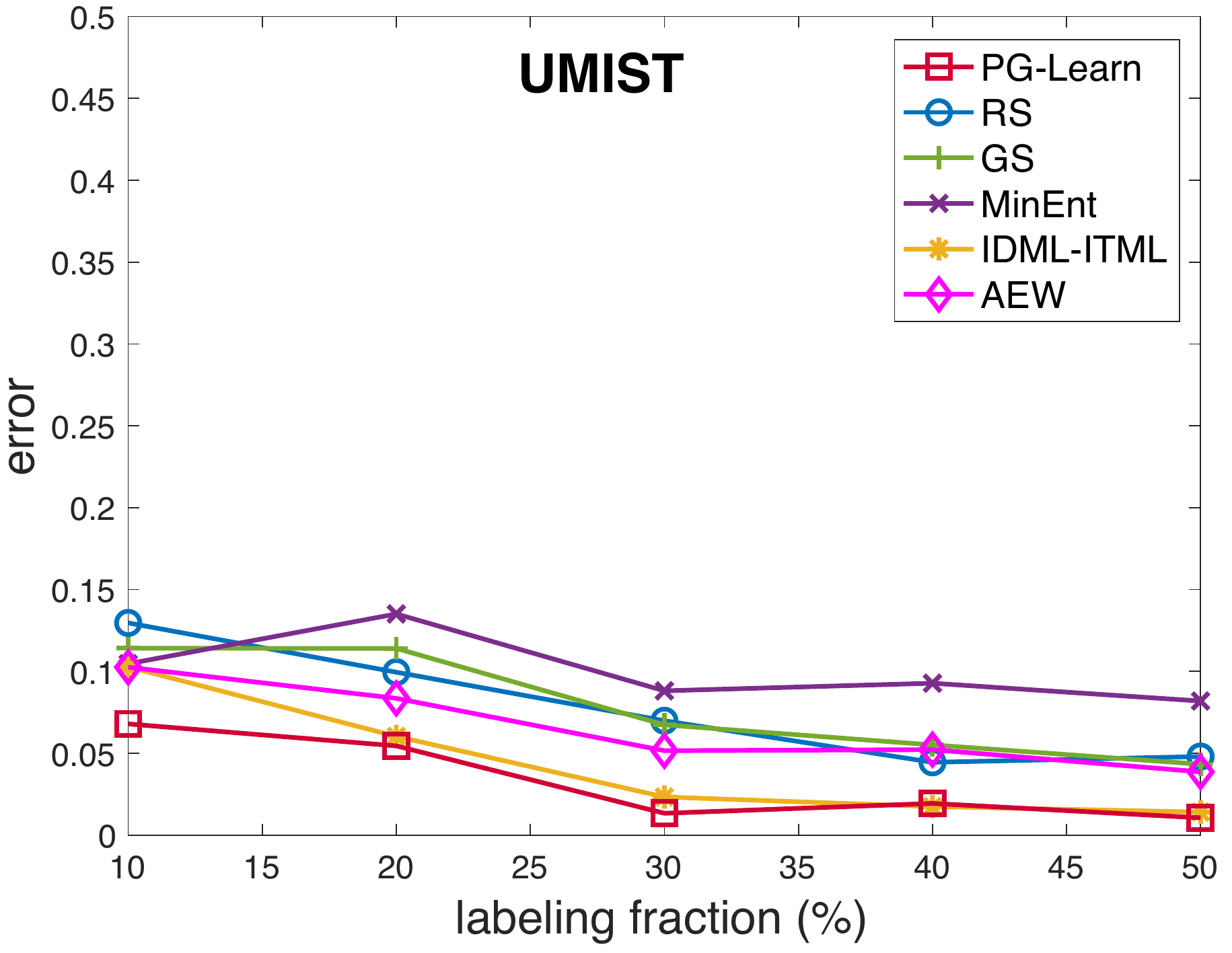}&
		\includegraphics[width=0.35\columnwidth]{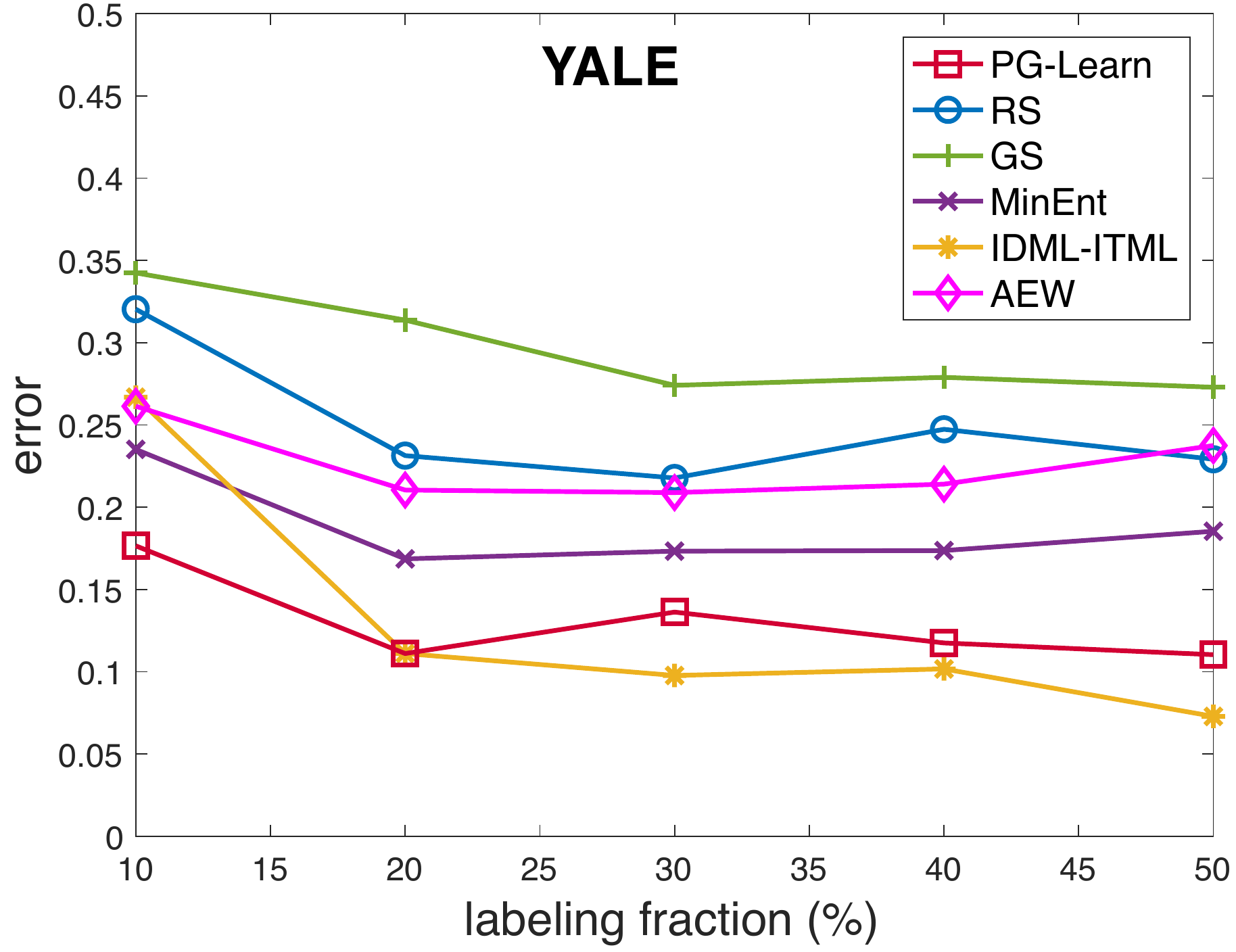}
	\end{tabular}
	 \vspace{-.1in}
	\caption{Test error (avg'ed across 3 random samples) as labeled data percentage is increased up to 50\%. \method~performs the best in many cases, and \textit{consistently} ranks in top two among competitors on each dataset and each labeling \%.}
	\label{fig:seq}
\end{figure}

Next we investigate how the prediction performance of the competing methods changes by varying labeling percentage.
To this end, we repeat the experiments using up to 50\% labeled data.
As shown in Figure \ref{fig:seq}, test error tends to drop with increasing amount of labels as expected. \method~ achieves the lowest error in many cases across datasets and labeling ratios. \minent~ is the closest competition on \usps~ and \mnist, which however ranks lower on \umist~and \yale. Similarly, \metric~is close competition on \umist~and \yale, which however performs poorly on \coil~ and \usps.
In contrast, \method~ {\em consistently} performs near the top.

\begin{table}[!t]
	\centering
	\caption{Test accuracy with 10\% labeled data, avg'ed across 10 random samples; 15 mins of hyperparameter tuning on single thread. Symbols $\blacktriangle$ ($p$$<$$0.005$) and $\triangle$ ($p$$<$$0.01$) denote the cases where \method\ is significantly better than the baseline w.r.t. the paired Wilcoxon signed rank test.
	} 
	\vspace{-0.1in}
	\scalebox{0.85}{
		\begin{tabular}{|l||l|l|l|l|l|l|} \toprule
			\textbf{Dataset}	&	\textbf{{\sc PG-Lrn}}		& \textbf{\minent}		&	\textbf{\metric}		&	\textbf{\aew}	&	\textbf{\grid}	&	\textbf{\randd} \\ 
			\midrule
			\coil & \textbf{0.9232} & 0.9116\s & 0.7508\s & 0.9100\s & 0.8929\s & 0.8764\s \\ \hline
			
			\usps & \textbf{0.9066} & \textbf{0.9088} & 0.8565\s & 0.8951\s & 0.8732\s & 0.8169\s \\ \hline
			\mnist & \textbf{0.8241} & \textbf{0.8163} & 0.7801\hs & 0.7828\s & 0.7550\s & 0.7324\s \\ \hline
			
			\umist & \textbf{0.9321} & 0.8954\s & 0.8973\hs & 0.8975\s & 0.8859\s & 0.8704\s \\ \hline
			
			\yale & \textbf{0.8234} & 0.7648\hs & 0.7331\s & 0.7386\s & 0.6576\s & 0.6797\s\\ 
			%
			\bottomrule			
		\end{tabular}}
		\label{tab:seq}
	\end{table}

We quantify the above more concretely, and provide  the test accuracy for each labeling \% in Table \ref{tab:percent}, averaged across random samples from all datasets, along with results of significance tests. We also 
give the average rank per method, as ranked by test error (hence, lower is better).

\method~ significantly outperforms \textit{all} competing methods in accuracy at \textit{all} labeling ratios w.r.t. the paired Wilcoxon signed rank test at $p=0.01$, as well as achieves the lowest rank w.r.t. test error.
On average, \minent~ is the closest competition, followed by \aew. Despite being supervised, \metric~ does not perform on par. This may be due to labeled data not being sufficient to learn a proper metric in high dimensions, and/or the labels introduced during self-learning being noisy. 
We also find \grid~ and \randd~ to rank at the bottom, suggesting that 
learning the graph structure provides advantage over
these standard techniques.

\begin{table}[!t]
	\centering
	\caption{Average test accuracy and rank (w.r.t. test error) of methods across datasets for varying labeling \%. $\blacktriangle$ ($p$$<$$0.005$) and $\triangle$ ($p$$<$$0.01$) denote the cases where \method\ is significantly better w.r.t. the paired Wilcoxon signed rank test.} 
	\vspace{-0.125in}
	\hspace{-0.075in}
	\scalebox{0.88}{
		\begin{tabular}{|r||c|c|c|c|c|c|} \toprule
			\textbf{Labeled}	&	\textbf{{\sc PG-L}}		& \textbf{\minent}		&	\textbf{\metric}		&	\textbf{\aew}	&	\textbf{\grid}	&	\textbf{\randd} \\ 
			\midrule
			10\% acc. & \textbf{0.8819} & 0.8594\s & 0.8036\s & 0.8448\s & 0.8129\s & 0.7952\s \\ 
			rank & {\textbf{1.20}} & 2.20 & 4.40 & 2.80 & 4.80  & 5.60 \\ \hline
			
			20\% acc. & \textbf{0.8900} & 0.8504\s & 0.8118\s & 0.8462\s & 0.8099\s & 0.8088\s \\ 
			rank & \textbf{1.42} & 2.83 & 4.17 & 2.92 & 4.83  & 4.83 \\ \hline
			30\% acc. & \textbf{0.9085} & 0.8636\s & 0.8551\s & 0.8613\s & 0.8454\s & 0.8386\s \\ 
			rank & \textbf{1.33} & 3.67 & 3.83 & 3.17 & 4.00  & 5.00 \\ \hline
			40\% acc. & \textbf{0.9153} & 0.8617\s & 0.8323\s & 0.8552\s & 0.8381\s & 0.8303\s \\ 
			rank & \textbf{1.67} & 3.67 & 3.50 & 3.67 & 4.00  & 4.50 \\ \hline
			50\% acc. & \textbf{0.9251} & 0.8700\hs & 0.8647\s & 0.8635\s & 0.8556\s & 0.8459\s \\ 
			rank & \textbf{1.50} & 3.17 & 3.83 & 3.67 & 4.00  & 4.83 \\ 
		 \bottomrule
		\end{tabular}}
		\label{tab:percent}
	\end{table}

\subsubsection{Parallel Experiments with Noisy Features} 
Next we fully evaluate \method~ in the parallel setting as proposed in Algo. \ref{alg:pglearn}.
Graph learning is especially beneficial for SSL in noisy scenarios, where there exist irrelevant or noisy features that would cause simple graph construction methods like $k$NN and \grid~go astray.
To the effect of making the classification tasks more challenging, we \textit{double} the feature space for each dataset, by adding 100\% new noise features with values drawn randomly from standard $Normal(0,1)$.
Moreover, this provides a ground truth on the importance of features, based on which
we are able to quantify how well our \method~ recovers the necessary underlying relations by learning the appropriate feature weights.

{\bf Setup:} We report results comparing \method~ only with \minent, \grid, and \randd---in this setup, \metric~ failed to learn a metric in several cases due to degeneracy and the authors' implementation\footref{note2} of \aew~ gave out-of-memory errors in many cases. This however does not take away much, since \minent~ proved to be the second-best after \method~ in the previous section (see Table \ref{tab:percent}) and \grid~ and \randd~ are the typical methods used often in practice.

Given a budget $B$ units of time and $T$ parallel threads for our \method, 
each competing method is executed for a total of $BT$ units, i.e. all methods receive the same amount of processing time.\footnote{All experiments executed on a Linux server equipped with 96 Intel Xeon CPUs at 2.1 GHz and a total of 1 TB RAM,
	using Matlab R2015b Distributed Computing Server.}
Specifically, \minent~ is started in $T$ threads, each with a random initial configuration that runs until time is up (i.e., to completion, no early-terminations). 
\grid~ picks $(k,\sigma)$ from the 2-d grid that we refine recursively, that is, split into finer resolution containing more cells as more allocated time remains, while
\randd~ continues picking random combinations of $(k,\ba_{1:d})$.
When the time is over, each method reports the hyperparameters that yield the highest validation accuracy, using which the test accuracy is computed.

{\bf Results:} 
Table \ref{tab:par} presents the average test accuracy over 10 random samples from each dataset, using $T=32$.
We find that despite $32\times$ more time, the baselines are crippled by the irrelevant features and increased dimensionality. In contrast, \method~ maintains notably high accuracy that is significantly better than all the baselines on all datasets at $p=0.01$.

\begin{figure}[!t]
	\centering
	\begin{tabular}{c|c}
		\hspace{-.1in}\includegraphics[width=0.475\columnwidth]{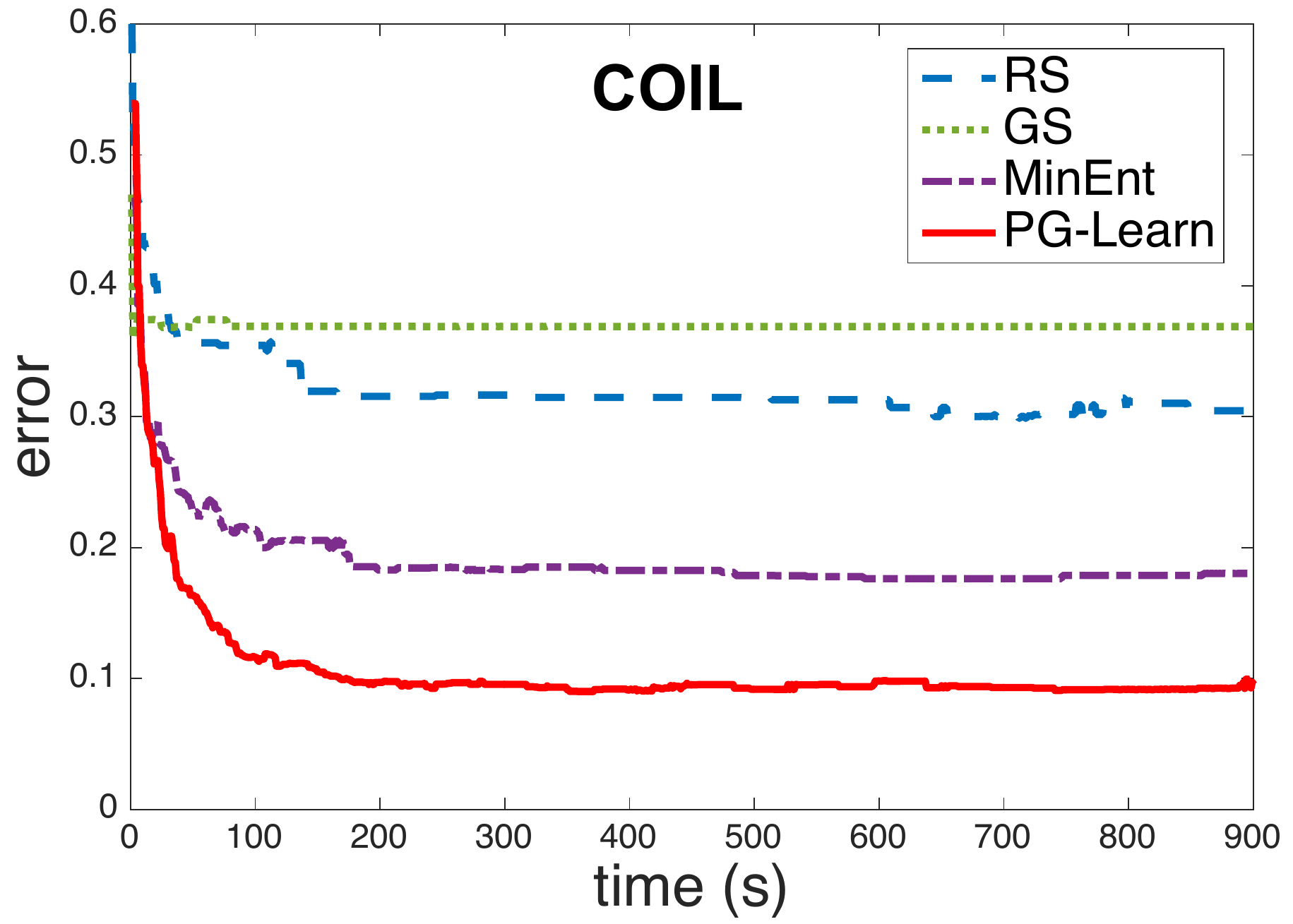}&
		\hspace{-.1in}\includegraphics[width=0.475\columnwidth]{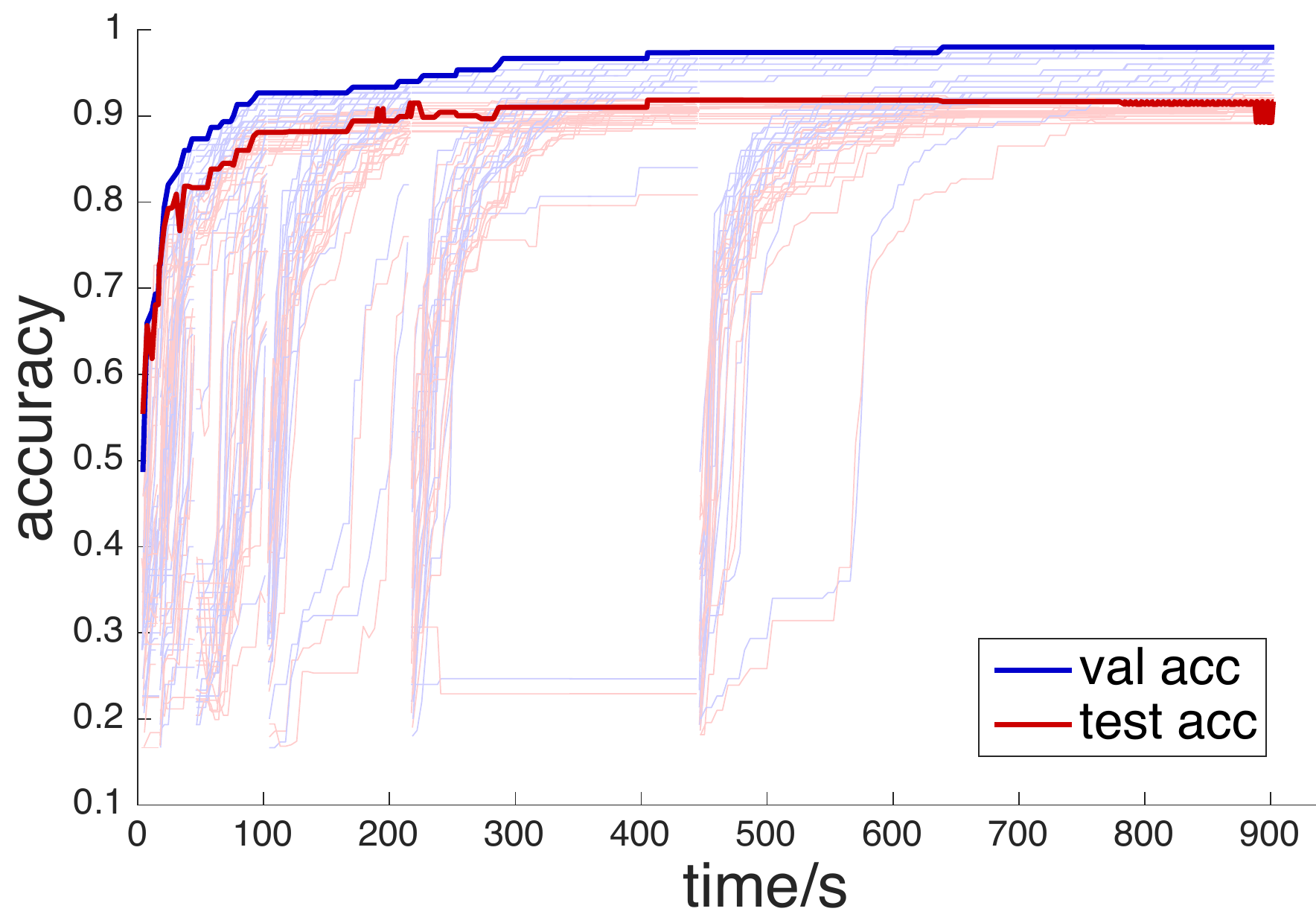}\\
		
		\hspace{-.1in}\includegraphics[width=0.475\columnwidth]{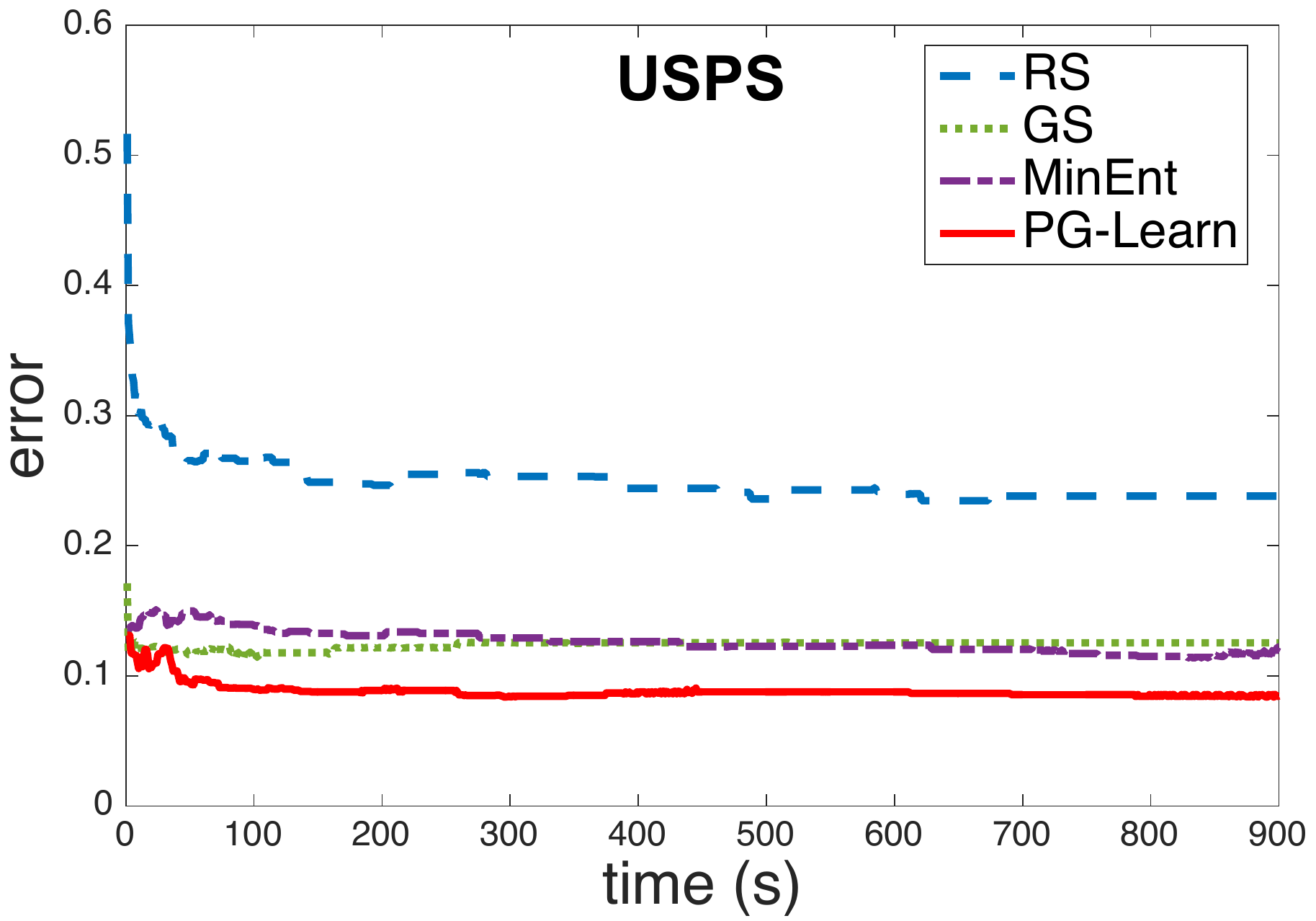}&
		\hspace{-.1in}\includegraphics[width=0.475\columnwidth]{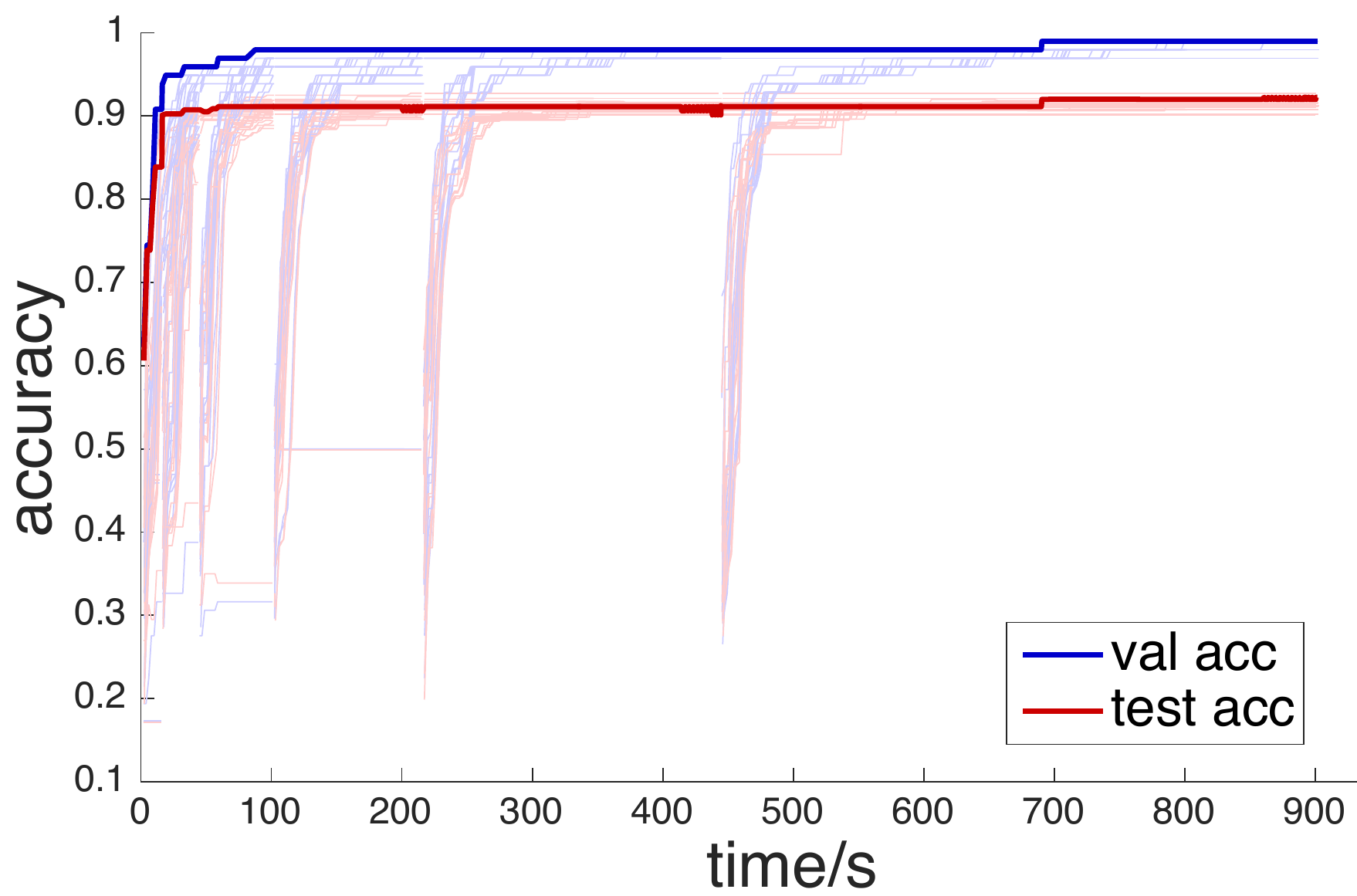}\\
		\hspace{-.1in}\includegraphics[width=0.475\columnwidth]{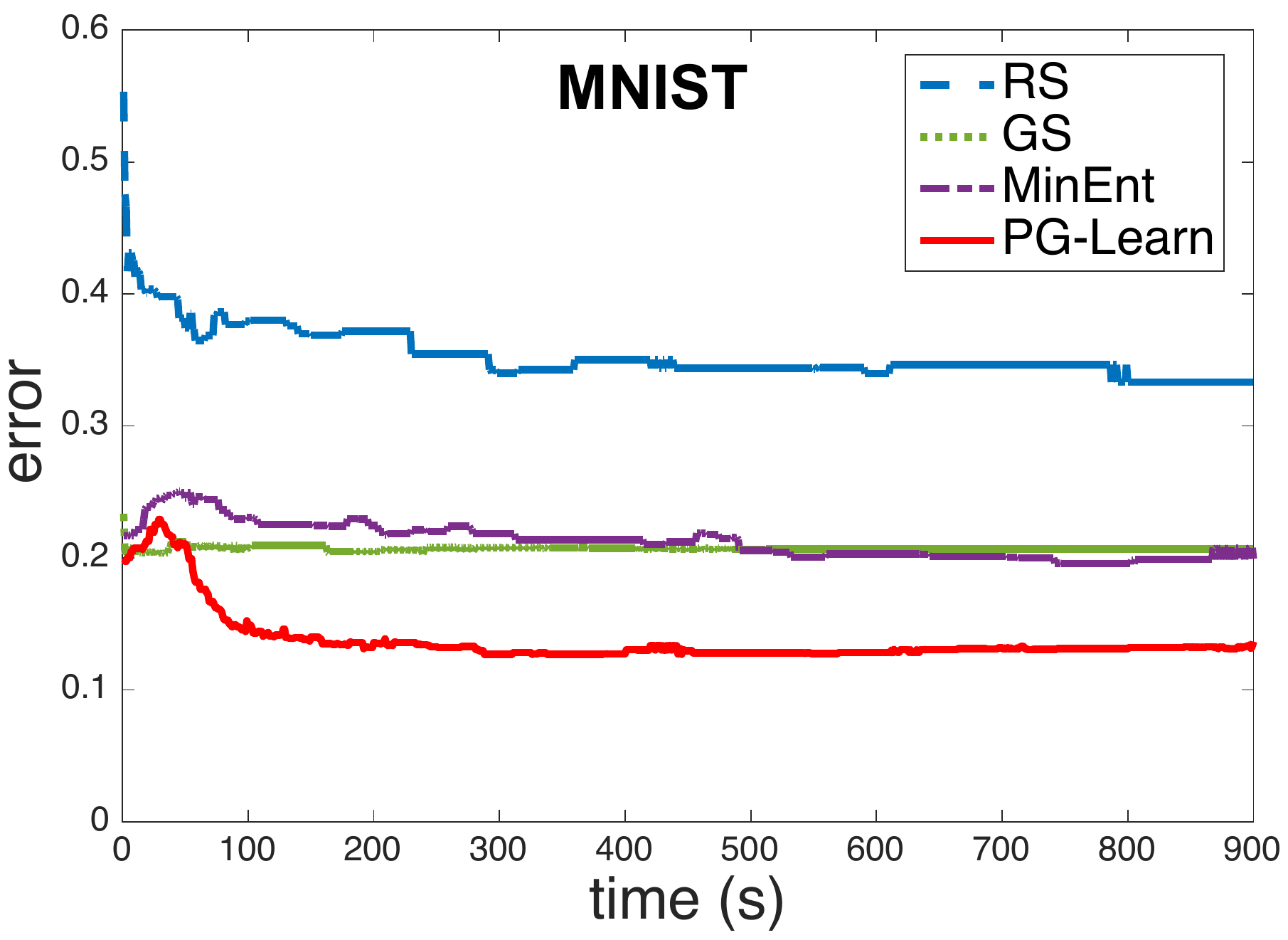}&
		\hspace{-.1in}\includegraphics[width=0.475\columnwidth]{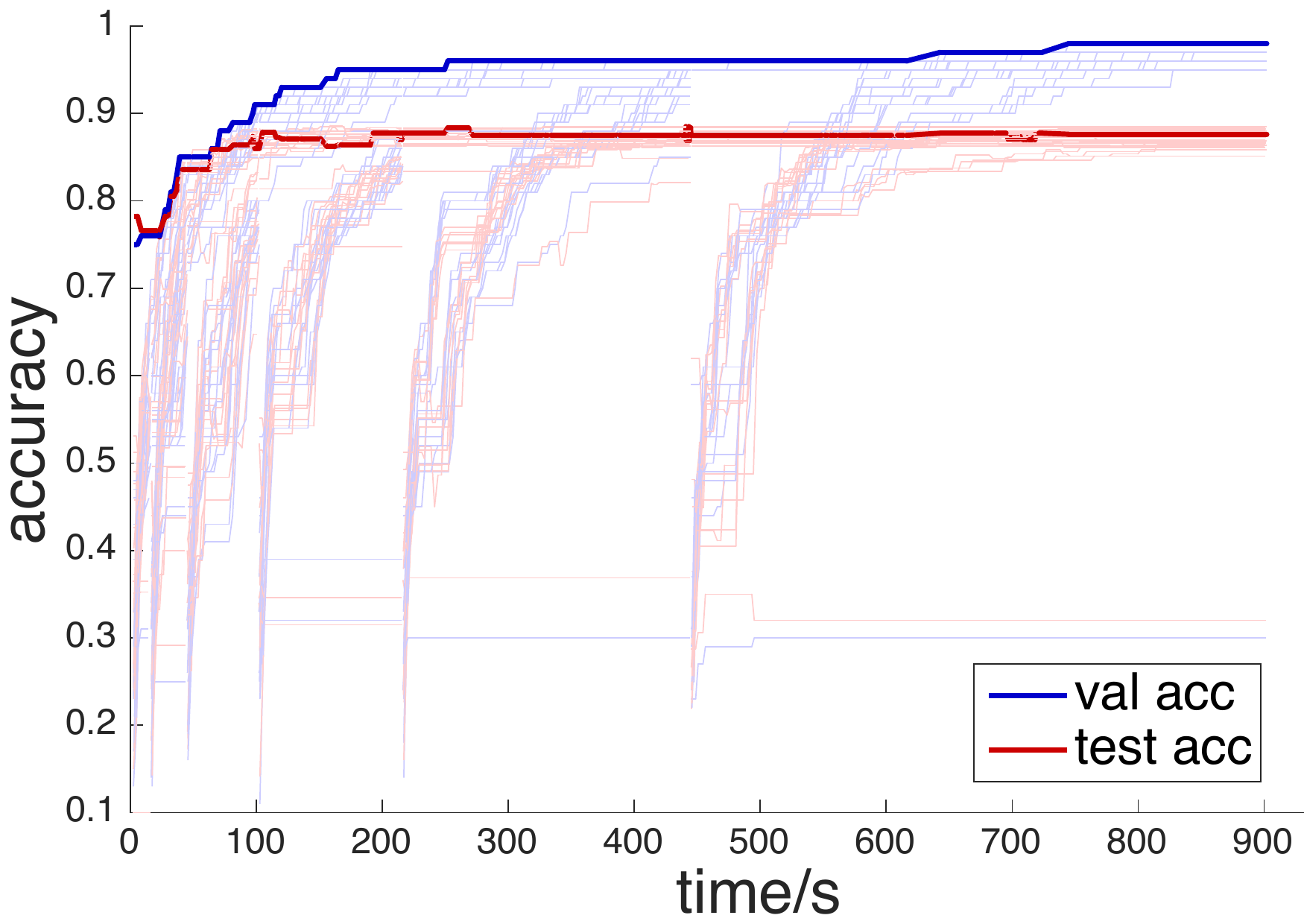}\\
		\hspace{-.1in}\includegraphics[width=0.475\columnwidth]{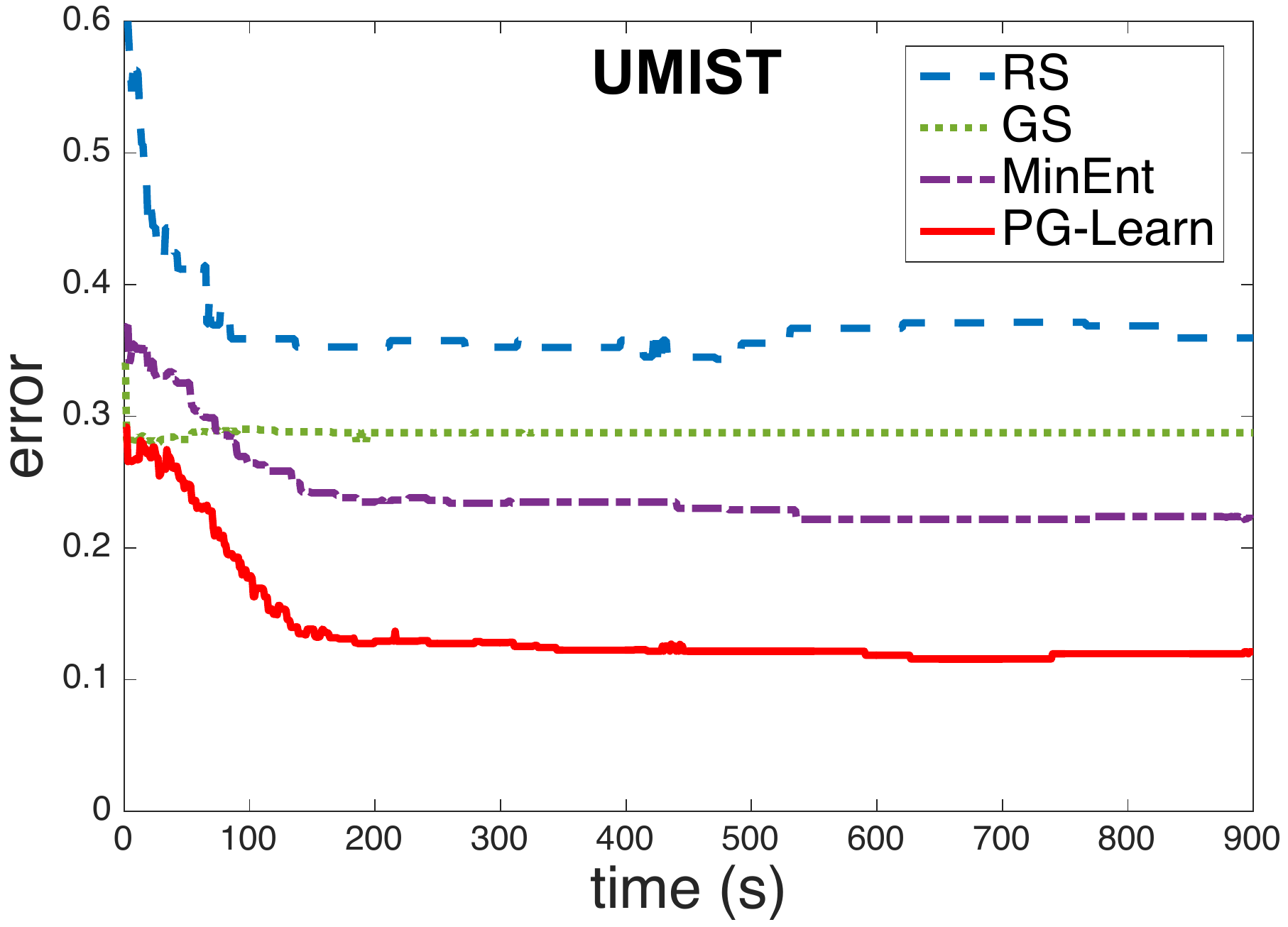}&
		\hspace{-.1in}\includegraphics[width=0.475\columnwidth]{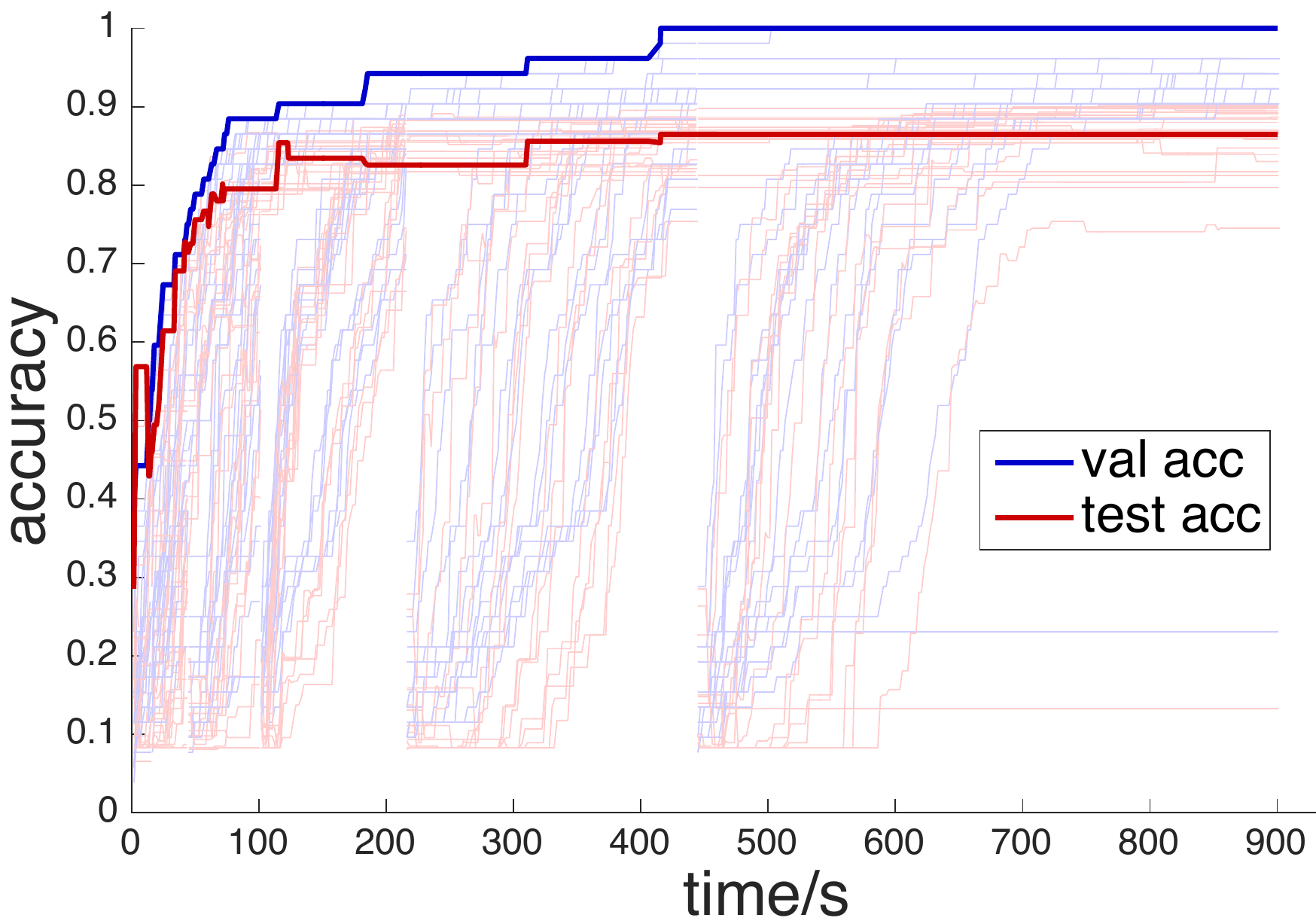}\\
		\hspace{-.1in}\includegraphics[width=0.475\columnwidth]{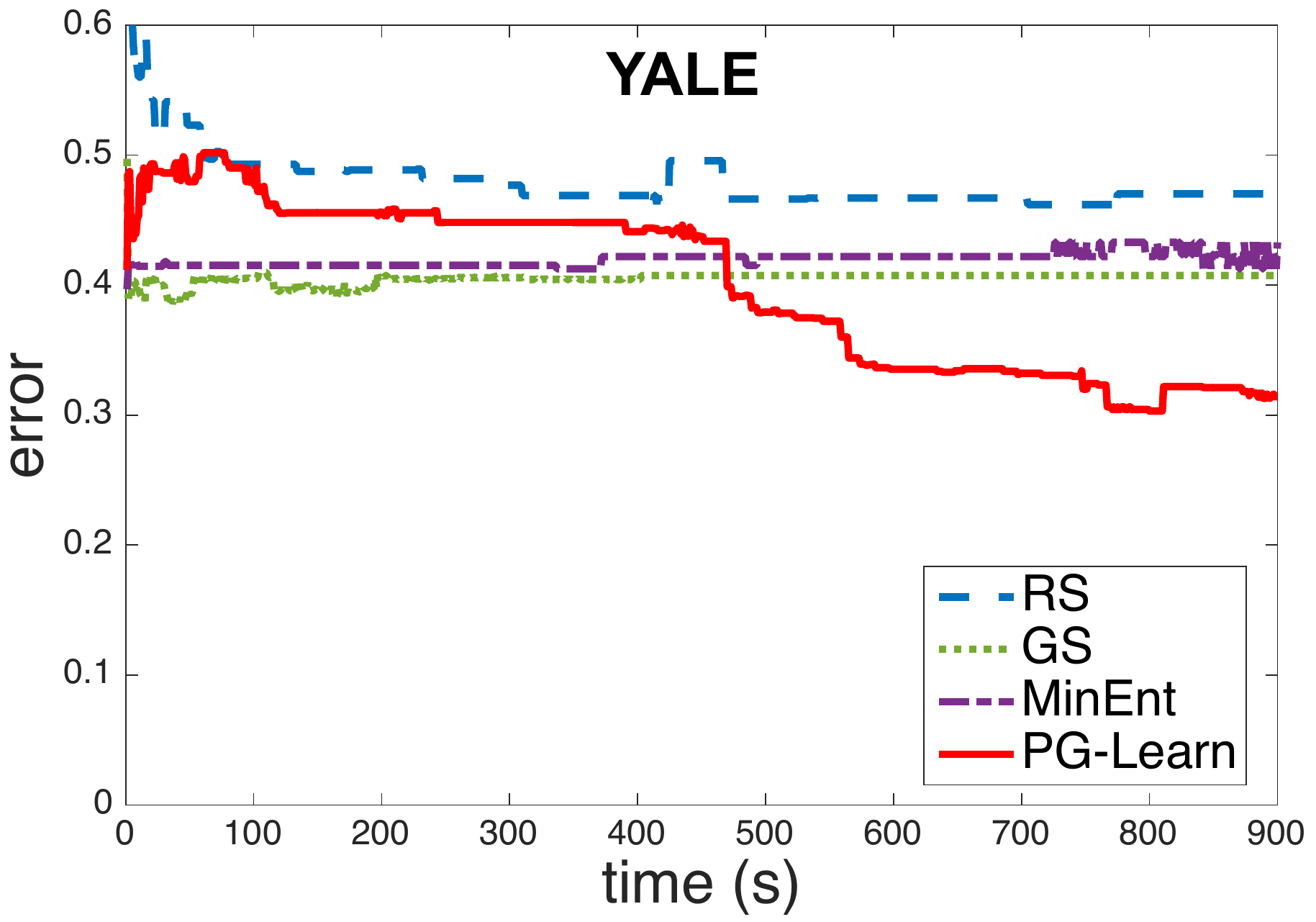}&
		\hspace{-.1in}\includegraphics[width=0.475\columnwidth]{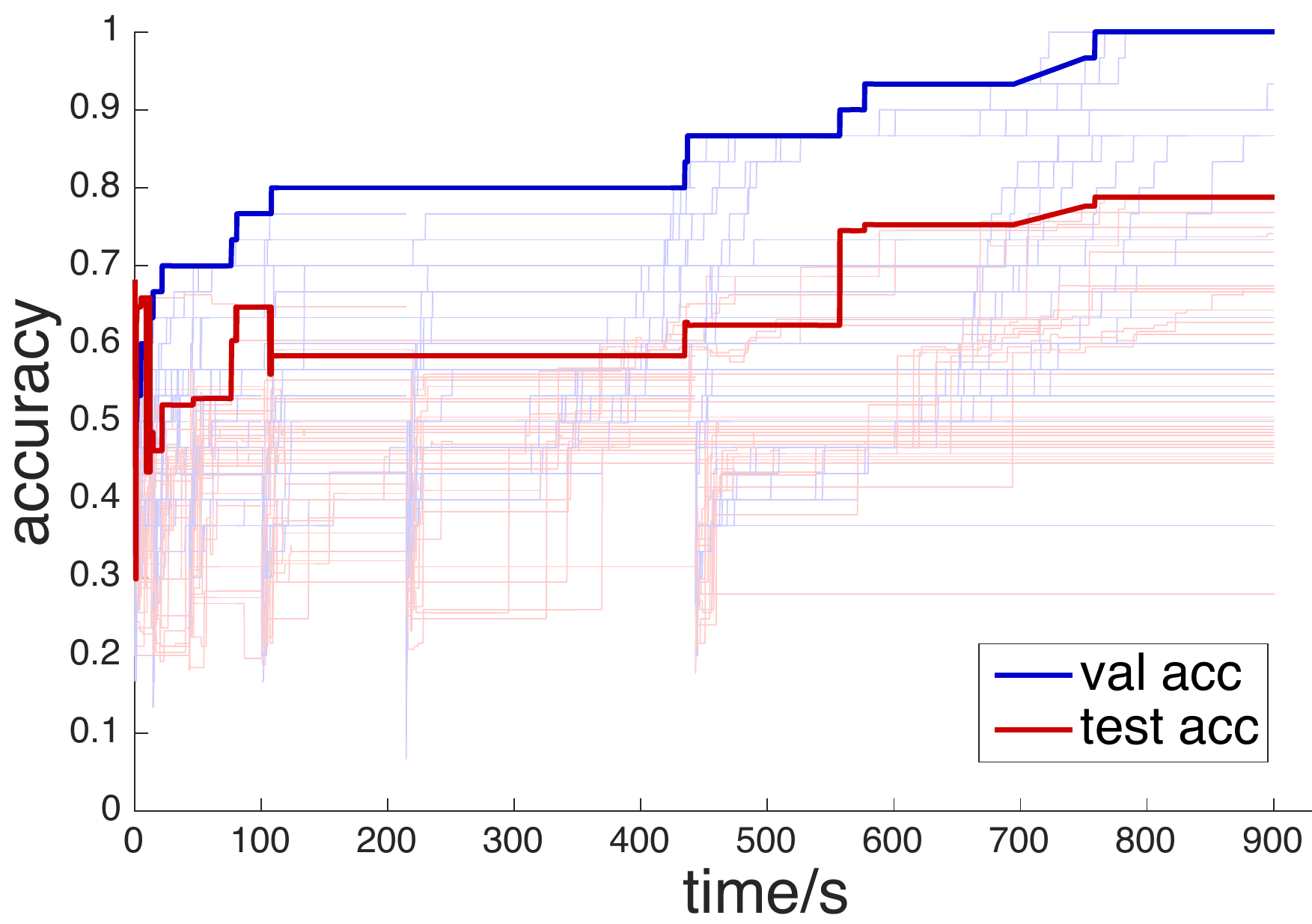}\\
		\small{\textbf{(a) test error by time}} & \small{\textbf{(b) \method~ val.\&test acc. by time}}  \\
	\end{tabular}
	\caption{(a) Test error vs. time (avg'ed across 10 runs w/ random samples) comparing \method~with baselines on noisy datasets; 
		(b) \method's validation and corresponding test accuracy over time as it executes Algo. \ref{alg:pglearn} on 32 threads (1 run).}
	\label{fig:testvstime}
\end{figure}

\begin{table}[!t]
	\centering
	\caption{Test accuracy on datasets with 100\% added noise features, avg'ed across 10 samples; 15 mins of hyperparameter tuning on $T=32$ threads. Symbols $\blacktriangle$ ($p$$<$$0.005$) and $\triangle$ ($p$$<$$0.01$) denote the cases where \method\ is significantly better than the baseline w.r.t. the paired Wilcoxon signed rank test.} 
	 \vspace{-0.125in}
	\scalebox{0.85}{
		\begin{tabular}{|l||l|l|l|l|} \toprule
			\textbf{Dataset}	&	\textbf{{\sc PG-Lrn}}		& \textbf{\minent}	&	\textbf{\grid}	&	\textbf{\randd} \\ 
			\midrule
			\coil & \textbf{0.9044} & 0.8197\s & 0.6311\s & 0.6954\s  \\ \hline
			
			\usps & \textbf{0.9154} & {0.8779}\hs & 0.8746\s & 0.7619\s  \\ \hline
			\mnist & \textbf{0.8634} & {0.8006}\s & 0.7932\s & 0.6668\s \\ \hline
			
			\umist & \textbf{0.8789} & 0.7756\s & 0.7124\s & 0.6405\s  \\ \hline
			\yale & \textbf{0.6859} & 0.5671\s & 0.5925\s & 0.5298\s \\ 
			%
			\bottomrule			
	\end{tabular}}
	\label{tab:par}
\end{table}
Figure \ref{fig:testvstime} (a) shows how the test error changes by time for all methods on average, and (b) depicts the validation and the corresponding test accuracies for \method~ on an example run.
We see that \method~ gradually improves validation accuracy across threads over time, and test accuracy follows closely. As such, test error drops in time.
\grid~search has a near-flat curve as it uses the same kernel bandwidth on all dimensions, therefore, more time does not help in handling noise.
\randd~ error seems to drop slightly but stabilizes at a high value, demonstrating its limited ability to guess parameters in high dimensions with noise. 
Overall, \method~ outperforms competition significantly in this high dimensional noisy setting as well.
Its performance is particularly noteworthy on \yale, which has small $n=320$ but large $2d>2K$ half of 
which are noise.

\begin{figure}[!t]
	\begin{tabular}{c} 
	\hspace{-0.1in}	\includegraphics[width=0.485\textwidth]{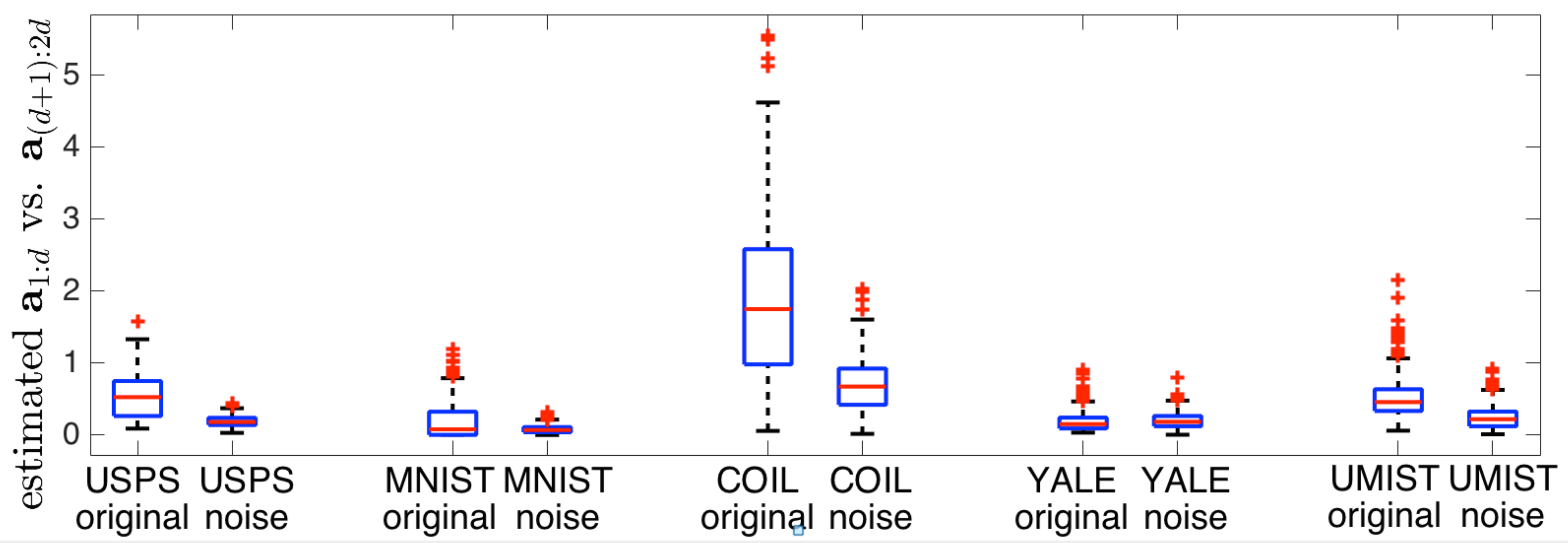} 
	\end{tabular}
 \vspace{-0.05in}
	\caption{Distribution of weights estimated by \method, shown separately for the original and injected noisy features on each dataset. Notice that the latter is much lower, showing it competitiveness in hyperparameter estimation.}
	\label{fig:weights}
\end{figure}

Finally, Figure \ref{fig:weights} shows \method's estimated hyperparameters, $\ba_{1:d}$ and $\ba_{(d+1):2d}$ (avg'ed over 10 samples), demonstrating that the noisy features $(d+1):2d$ receive notably lower weights.

%% file: 05conclusion.tex
In this work we addressed the graph structure estimation problem as part of relational semi-supervised inference.
It is now well-understood that graph construction from point-cloud data has critical impact on learning algorithms \cite{conf/nips/MaierLH08,conf/pkdd/SousaRB13}.
To this end, we first proposed a learning-to-rank based objective parameterized by different weights per dimension and derived its gradient-based learning (\S\ref{ssec:gloss}).
We then showed how to integrate this type of adaptive local search within a parallel framework that early-terminates searches based on relative performance, in order to dynamically allocate resources (time and processors) to those with promising configurations (\S\ref{ssec:parallel}).
Put together, our solution \method~ is a hybrid that strategically navigates the hyperparameter search space.
What is more, \method~ is scalable in dimensionality and number of samples both in terms of runtime and memory requirements.


As future work we plan to deploy \method~ on a distributed platform like Apache Spark, and generalize the ideas to other graph-based learning problems such as graph-regularized regression.